\documentclass[3p,twocolumn,preprint]{elsarticle}

\usepackage{lineno,hyperref}
\usepackage{algorithm}
\usepackage{mathtools}
\usepackage{amssymb}
\usepackage{url}
\usepackage{amsmath}
\usepackage{graphicx}
\usepackage{subcaption}
\usepackage{array}
\usepackage{algorithm}
\usepackage[noend]{algpseudocode}
\usepackage[misc]{ifsym}
\usepackage{fixltx2e}
\usepackage{tikz}
\usepackage{color}
\usetikzlibrary{positioning}
\usepackage{float}

\modulolinenumbers[5]

\journal{NeuroImage}

\begin{document}

\begin{frontmatter}

\title{Disease Progression Timeline Estimation for Alzheimer's Disease using Discriminative Event Based Modeling}

\author[add1]{Vikram Venkatraghavan\corref{mycorrespondingauthor}}
\cortext[mycorrespondingauthor]{Corresponding author}
\ead{v.venkatraghavan@erasmusmc.nl}
\author[add1]{Esther E. Bron}
\author[add1,add2]{Wiro J. Niessen}
\author[add1]{Stefan Klein}
\author{for the Alzheimer's Disease Neuroimaging Initiative \fnref{fn1}}
\fntext[fn1]{Data used in preparation of this article were obtained from the Alzheimer's Disease Neuroimaging Initiative (ADNI) database (adni.loni.usc.edu). As such, the investigators within the ADNI contributed to the design and implementation of ADNI and/or provided data but did not participate in analysis or writing of this report. A complete listing of ADNI investigators can be found at: \url{http://adni.loni.usc.edu/wp-content/uploads/how_to_apply/ADNI_Acknowledgement_List.pdf}}

\address[add1]{Biomedical Imaging Group Rotterdam, Departments of Medical Informatics \& Radiology, Erasmus MC, University Medical Center Rotterdam, The Netherlands}
\address[add2]{Quantitative Imaging Group, Dept. of Imaging Physics, Faculty of Applied Sciences, Delft University of Technology, Delft, The Netherlands}

\begin{abstract}
Alzheimer's Disease (AD) is characterized by a cascade of biomarkers becoming abnormal, the pathophysiology of which is very complex and largely unknown. Event-based modeling (EBM) is a data-driven technique to estimate the sequence in which biomarkers for a disease become abnormal based on cross-sectional data. It can help in understanding the dynamics of disease progression and facilitate early diagnosis and prognosis by staging patients. In this work we propose a novel discriminative approach to EBM, which is shown to be more accurate than existing state-of-the-art EBM methods. The method first estimates for each subject an approximate ordering of events. Subsequently, the central ordering over all subjects is estimated by fitting a generalized Mallows model to these approximate subject-specific orderings based on a novel probabilistic Kendall's Tau distance. We also introduce the concept of relative distance between events which helps in creating a disease progression timeline. Subsequently, we propose a method to stage subjects by placing them on the estimated disease progression timeline. We evaluated the proposed method on Alzheimer's Disease Neuroimaging Initiative (ADNI) data and compared the results with existing state-of-the-art EBM methods. We also performed extensive experiments on synthetic data simulating the progression of Alzheimer's disease. The event orderings obtained on ADNI data seem plausible and are in agreement with the current understanding of progression of AD. The proposed patient staging algorithm performed consistently better than that of state-of-the-art EBM methods. Event orderings obtained in simulation experiments were more accurate than those of other EBM methods and the estimated disease progression timeline was observed to correlate with the timeline of actual disease progression. The results of these experiments are encouraging and suggest that discriminative EBM is a promising approach to disease progression modeling.

\end{abstract}


\end{frontmatter}

\nolinenumbers

\section{Introduction} \label{sec:intro}

Dementia is considered a major global health problem as the number of people living with dementia was estimated to be about $46.8$ million in $2015$. It is expected to increase to $131.5$ million in $2050$~\citep{Prince:2015}. Alzheimer's Disease (AD) is the most common form of dementia. There is a gradual shift in the definition of AD from it being a clinical-pathologic entity (based on clinical symptoms), to a biological one based on neuropathologic change (change of imaging and non-imaging biomarkers from normal to abnormal)~\citep{Jack:2018}. The latter definition is more useful for understanding the mechanisms of disease progression.

Preventive and supportive therapy for patients at risk of developing dementia due to AD could improve their quality of life and reduce costs related to care and lifestyle changes. To identify the at-risk individuals as well as monitor the effectiveness of these preventive and supportive therapies, methods for accurate patient staging (estimating the disease severity in each individual) are needed. To enable accurate patient staging in an objective and quantitative way, it is important to understand how the different imaging and non-imaging biomarkers progress after disease onset.

Longitudinal models of disease progression reconstruct biomarker trajectories in individual subjects~\citep{Donohue:2014, Sabuncu:2014, Schmidt:2016}. However, the utility of such models is restricted by the fact that longitudinal data in large groups of patients is scarce. Identifying at-risk individuals for clinical trials by studying their biomarker trajectory evolution is also not feasible. 

\begin{figure}[t!]
\centering
\includegraphics[width=0.40\textwidth]{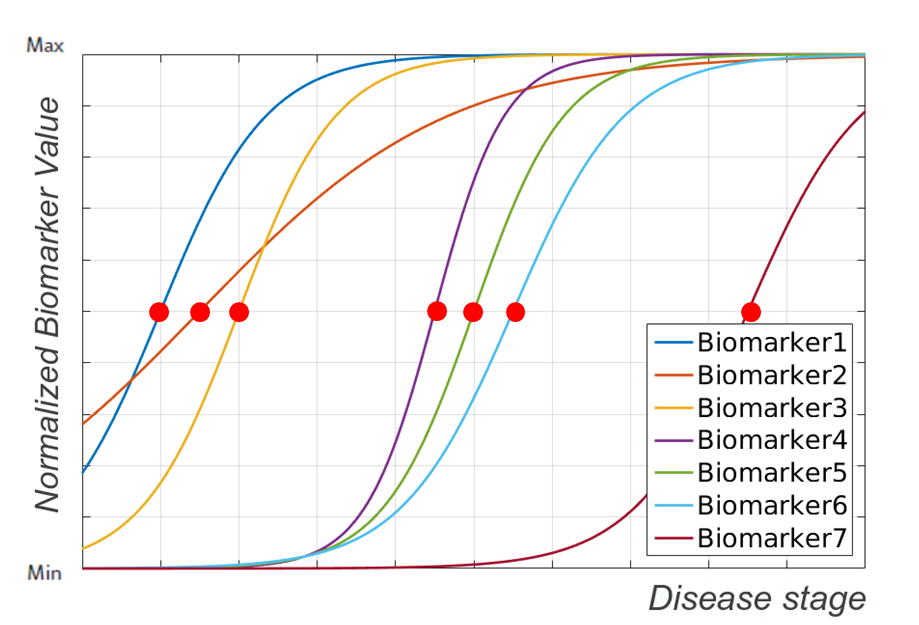}
\caption{Illustration of the output expected in an EBM. The biomarker
trajectories shown here are hypothetical trajectories representing a change of biomarker value from normal state. The dots on these trajectories are biomarker events as defined in an EBM. Output of an EBM is the ordering of such events.}
\label{fig:EBM-Illu}
\end{figure}

To circumvent this problem, methods to infer the order in which biomarkers become abnormal during disease progression based on cross-sectional data have been proposed~\citep{Fonteijn:2012, Huang:2012, Medina:2016}. The model used in~\cite{Medina:2016} relies on stratification of patients into several subgroups based on symptomatic staging, for inferring the aforementioned ordering. However, the problem with using symptomatic staging is that it is very coarse and qualitative. The models used in~\cite{Fonteijn:2012, Huang:2012} are variants of Event-Based Models (EBM). EBM algorithms neither rely on symptomatic staging nor on the presence of longitudinal data for inferring the temporal ordering of events, where an event is defined by a biomarker becoming abnormal. Figure~\ref{fig:EBM-Illu} shows these biomarker events on hypothetical trajectories as expected in a typical neuropathologic change. 

An important assumption made in~\cite{Fonteijn:2012} is that the ordering of events is common for all the subjects in a dataset. AD is known to be a heterogeneous disease with multiple disease subtypes. The assumptions in Fonteijn's EBM may therefore be too restrictive. The assumptions in Huang's EBM on the other hand are more realistic, as they do assume that the disease is heterogeneous. However the algorithm does not scale well to a large number of biomarkers~\citep{Vikram:2017}. 

To make EBM more scalable to large number of biomarkers and subjects, as well as make it robust to variations in ordering, we propose a novel approach to EBM, discriminative event-based model (DEBM), for estimating the ordering of events\footnote{An earlier version of the model was presented at the IPMI conference~\citep{Vikram:2017}. In the current manuscript, several methodological improvements and extensions are presented, and the experimental evaluation has been expanded substantially.}. We also introduce the concept of relative distance between events which helps in creating a disease progression timeline. Subsequently, we propose a method to stage subjects by placing them on the estimated disease progression timeline. The other contributions of this paper include an optimization technique for Gaussian mixture modeling that helps in accurate estimation of event ordering in DEBM as well as improving the accuracies of other EBMs, and a novel probabilistic distance metric between event orderings (probabilistic Kendall's Tau).

The remainder of the paper is organized as follows: An introduction to the existing EBM models is given in Section~\ref{sec:relatedwork}. In Section~\ref{sec:DEBM}, we propose our novel method for estimating central ordering of events. We perform extensive sets of experiments on ADNI data as well as on simulation data, the details of which are in Section~\ref{sec:exp}. Section~\ref{sec:results} summarizes the results of the experiments. Section~\ref{sec:disc} discusses the implications of these findings followed by concluding remarks in Section~\ref{sec:conc}.

\section{Event-Based Models} \label{sec:relatedwork}

EBM assumes monotonic increase or decrease of biomarker values with increase in disease severity (with the exception of measurement noise). It considers disease progression as a series of events,
where each event corresponds to a new biomarker becoming abnormal. Fonteijn's EBM~\citep{Fonteijn:2012} finds the ordering of events $(S)$ such that the likelihood that a dataset was generated from subjects following this event ordering is maximized. $S$ is a set of integer indices of biomarkers, which represents the order in which they become abnormal. Thus, disease progression is defined by $\{E_{S(1)}, E_{S(2)}, ..., E_{S(N)}\}$, where $N$ is the number of biomarkers per subject in the dataset and $E_{S(i)}$ is the $i$-th event that is associated with biomarker $S(i)$ becoming abnormal.

In a cross-sectional dataset $(X)$ of $M$ subjects, $X_j$ denotes a measurement of biomarkers for subject $j\in\left[1,M\right]$, consisting of $N$ scalar biomarker values $x_{j,i}$. Probabilistic formulation of an EBM, as proposed in~\cite{Fonteijn:2012}, can be given by $argmax_S(p(S|X))$, where $p(S|X)$ can be written using Bayes' rule as:

\begin{equation}
\label{eq:font0}
p(S|X) = \frac{p(S)p(X|S)}{p(X)}
\end{equation}

\noindent An important assumption in~\cite{Fonteijn:2012} is that $p(S)$ is uniformly distributed. This makes inferring $S$, equivalent to the maximum likelihood problem of maximizing $p\left( X|S \right)$\footnote{Fonteijn's EBM uses Markov Chain Monte Carlo (MCMC) sampling to estimate the posterior distribution $P(S|X)$. Average position of events in all the MCMC samples was used as a way for selecting the mean ordering by ~\cite{Fonteijn:2012} whereas further extensions of the work such as ~\cite{Young:2014} prefer the maximum likelihood solution.} . This can be further written in terms of $X_j$ as follows:

\begin{equation}
\label{eq:font1}
p\left( X|S \right) = \prod_{j=1}^{M} p\left( X_j|S \right)
\end{equation}

\noindent where $p\left( X_j|S \right)$ can be written as:

\begin{equation}
\label{eq:font112}
p\left( X_j|S \right) = \sum_{k=0}^N p(k|S)  p\left(X_j|k,S \right)
\end{equation}

\noindent where $p(k|S)$ is the prior probability of a subject being at position $k$ of the event ordering, which is assumed to be equal for each position. The $k$ which maximizes $p\left( X_j|S \right)$ denotes subject $j$'s disease stage. This method of identifying disease severity for a subject results in discrete set of stages, where the number of stages is one more than the number of biomarkers used for creating the model. $p\left(X_j| k,S\right)$ can be expressed as:

\begin{multline}
\label{eq:font12}
p\left(X_j|k, S \right) = \prod_{i=1}^k p \left( x_{j,S(i)} | E_{S(i)} \right) \times \\ \prod_{i=k+1}^N p \left( x_{j,S(i)} | \neg E_{S(i)} \right)
\end{multline}

\noindent where $p \left( x_{j,S(i)} | E_{S(i)} \right)$ is the likelihood of observing $x_{j,S(i)}$ in subject $j$, conditioned on event $i$ having already occurred. $p \left( x_{j,S(i)} | \neg E_{S(i)} \right)$, on the other hand, computes a similar likelihood, given that event $i$ has not occurred.

With the assumption that all the biomarkers in the control population are normal and that the biomarker values follow a Gaussian distribution, $p \left( x_{j,S(i)} | \neg E_{S(i)} \right)$ is computed. Abnormal biomarker values in the patient population are assumed to follow a uniform distribution but not all biomarkers of a patient could be assumed to be abnormal. For this reason, the likelihoods were obtained using a mixture model of a Gaussian and a uniform distribution, where only the parameters of the uniform distribution were allowed to be optimized.

This method was modified in~\cite{Young:2014} to estimate the optimal ordering in a sporadic AD dataset with significant proportions of controls expected to have presymptomatic AD~\citep{Schott:2010}. A Gaussian distribution was used to describe both the control and patient population, and the mixture model allowed for optimization of parameters for the Gaussians describing both control and patient population. The Gaussian mixture model was also used to incorporate more subjects from the dataset with clinical diagnosis of mild cognitive impairment (MCI).

After obtaining the central ordering $S$ which maximizes the likelihood $p\left( X|S \right)$, staging of patients is done by finding a disease stage $k$ for subject $j$, such that $p\left( X_j |k, S \right)$ is maximized.

The assumption that subjects follow a unique event ordering was relaxed by~\cite{Huang:2012}, who estimate a distribution of event orderings with a central event ordering $(S)$ and a spread $(\phi)$ as per a generalized Mallows model~\citep{Fligner:1988} using an expectation maximization algorithm. The E-step estimates the likelihood of patients' biomarker value measurements following subject-specific event order $s_j$, given $S$ and $\phi$. In the M-step, $S$ and $\phi$ are estimated based on $s_j$ estimated in the E-step. This is done iteratively to maximize the likelihood of generation of patients' data based on $S$ and $\phi$. Patient staging in Huang's EBM is also a maximum likelihood estimate, but unlike Fonteijn's EBM, the staging is done on the subject-specific event ordering $s_j$.

In both Fonteijn's and Huang's EBM, relative distances between events, that can be observed in Figure~\ref{fig:EBM-Illu}, are not captured\footnote{~\cite{Fonteijn:2012} briefly mention the idea of capturing relative distance between events, but it was not validated or used in any of the experiments.}. Some events can be closer to each other than others and using these relative distance between events could help create a more informative disease progression model.

\section{Discriminative Event-Based Model} \label{sec:DEBM}

\begin{figure}[t!]
\centering
\includegraphics[width=0.45\textwidth]{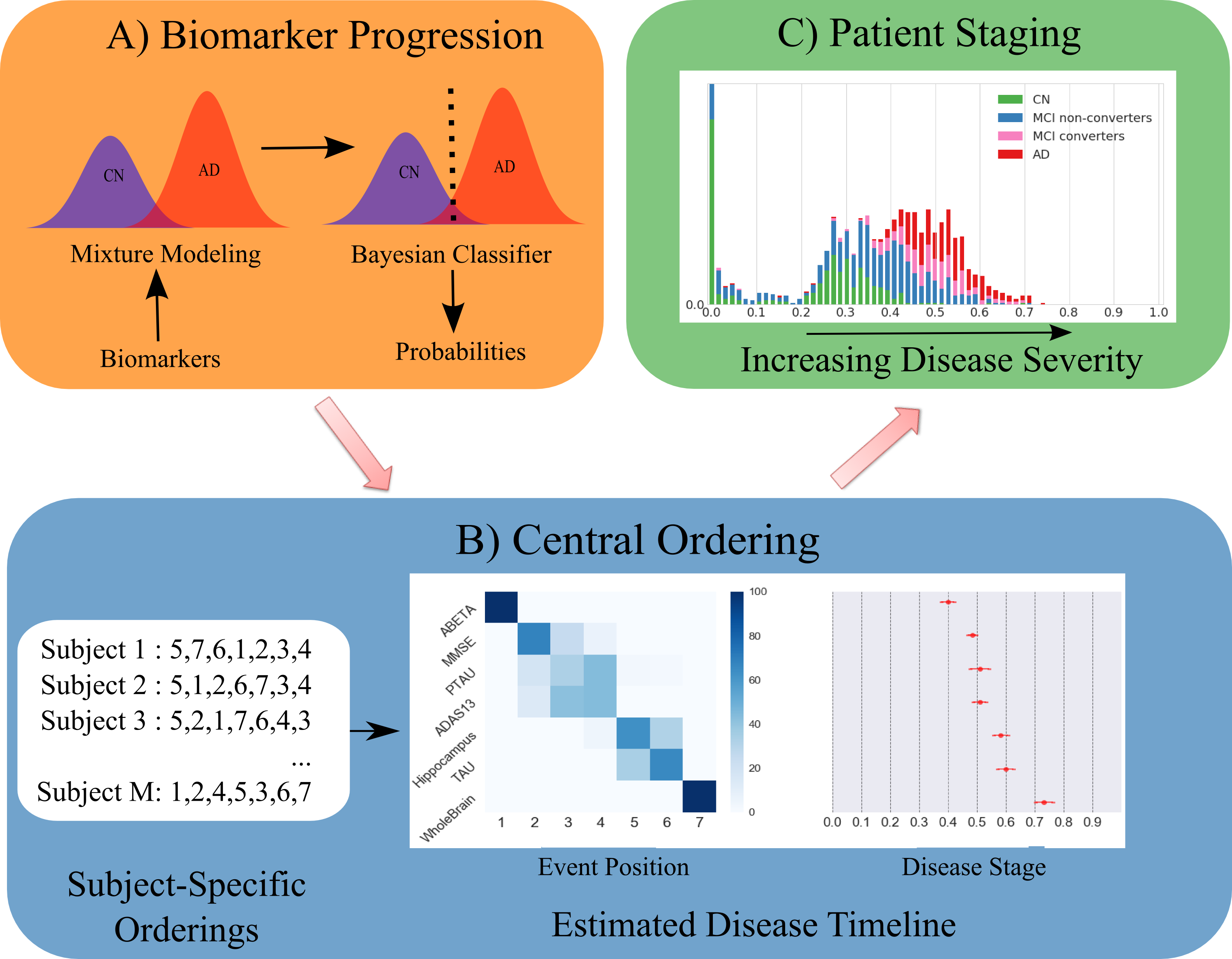}
\caption{Overview of the steps in DEBM. A) Biomarkers measured from different subjects are converted to probabilities of abnormality for individual biomarkers. This is done by estimating normal and abnormal distributions using Gaussian mixture modeling before classifying individual biomarkers using a Bayesian classifier. B) Subject-specific orderings of biomarker abnormalities are inferred from these probabilities which are then used to estimate the central ordering and for creating the disease progression timeline. C) This is then used to stage subjects based on disease severity.}
\label{fig:DEBM-Illu}
\end{figure}

Fonteijn's and Huang's EBM are generative models where the likelihood $p\left( X|S \right)$ is maximized. Huang's EBM also estimates subject-specific ordering based on a generative approach. Here, we propose our novel method for estimating central ordering of events $(S)$, a discriminative event-based model (DEBM). 

The proposed framework is discriminative in nature, since we estimate $s_j$ directly based on the posterior probabilities of individual biomarkers becoming abnormal. We also introduce a new concept of relative distance between events. This subsequently leads to a novel continuous patient staging algorithm. Figure~\ref{fig:DEBM-Illu} shows the different steps involved in our approach.

In Section~\ref{ssec:subjectordering}, we present the method to robustly estimate biomarker distributions in pre-event and post-event classes, given a single cross-sectional measurement of biomarkers. In Section~\ref{ssec:CentralOrdering}, we present a way for estimating $s_j$, and we address the problem of estimating a disease timeline from noisy estimates of $s_j$. In Section~\ref{ssec:PatientStaging}, we present the continuous patient staging method.

\subsection{Biomarker Progression} \label{ssec:subjectordering}

In this section, we propose a method to robustly convert $x_{j,i}$ to $p \left( E_{i} | x_{j,i} \right)$, which denotes the posterior probability of a biomarker measurement being abnormal. Assuming a paradigm similar to that in previous EBM variants~\citep{Huang:2012, Young:2014}, the probability density functions (PDF) of pre-event ($p \left( x_{j,i} | \neg E_{i} \right)$) and post-event ($p \left( x_{j,i} | E_{i} \right)$) classes in the biomarkers are assumed to be represented by Gaussians, independently for each biomarker. There are two reasons why constructing these PDFs is non-trivial. Firstly, the labels (clinical diagnoses) for the subjects do not necessarily represent the true labels of all the biomarkers extracted from the subject. Not all biomarkers are abnormal for subjects with AD diagnosis, while some of the cognitively normal (CN) subjects could have undiagnosed pre-symptomatic conditions. Secondly, the clinical diagnosis can be non-binary and include classes such as MCI, with significant number of biomarkers in normal and abnormal classes.

In our approach we address these two issues independently. We make an initial estimate of the PDFs using biomarkers from easily classifiable CN and easily classifiable AD subjects and later refine the estimated PDF using the entire dataset.

\begin{figure}[t!]
\centering
\includegraphics[width=0.48\textwidth]{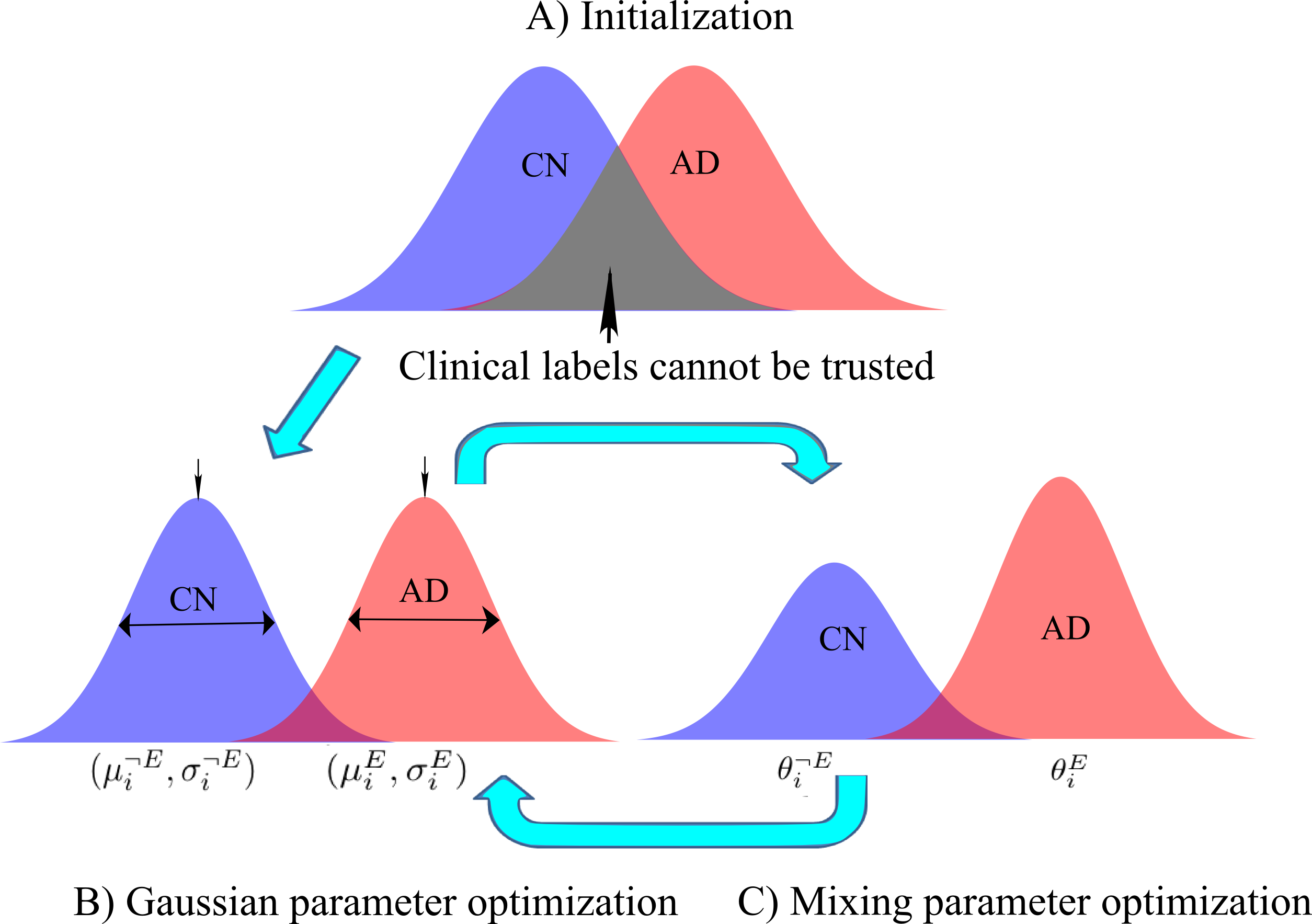}
\caption{Overview of the steps involved in the proposed Gaussian Mixture Model optimization strategy. A) Illustration of the initialization step for Gaussian Mixture Model. Rejecting the tails of the Gaussian distribution in CN and AD class is done to account for the fact that some of the CN subjects could be in pre-symptomatic stage of disease progression and some of the biomarkers could still be normal in AD subjects. B and C) This is followed by iterative estimation of Gaussian parameter optimization and Mixing parameter optimization.}
\label{fig:GMM_Illu}
\end{figure}

A Bayesian classifier is trained for each biomarker using CN and AD subjects, based on the assumption that there are no biomarkers in the pre-symptomatic stage for CN subjects and all the biomarkers are abnormal for AD subjects. This classifier is subsequently applied to the training data, and the predicted labels are compared with the clinical labels. The misclassified data in the dataset could either be outliers in each class resulting from our aforementioned assumption or could genuinely belong to their respective classes and represent the tails of the true PDFs. Irrespective of the reason of misclassification, we remove them for initial estimation of the PDFs. This procedure thus results, for each biomarker, in a set of easily classifiable CN subjects (whose biomarker values represent normal values) and easily classifiable AD subjects (whose biomarker values represent abnormal values). This is shown in the top part of Figure~\ref{fig:GMM_Illu}.

As we use Gaussians to represent the PDFs, we calculate initial estimates for mean and standard deviation for both normal $(\mu_{i}^{\neg E},\sigma_{i}^{\neg E})$ and abnormal classes $(\mu_{i}^E,\sigma_{i}^E)$ based on `easy' CN and `easy' AD subjects for each biomarker $i$. As these means and standard deviations are estimated based on truncated Gaussians, these are biased estimates. The initial estimates of standard deviations are always smaller than the expected unbiased estimates whereas the initial estimates of means are underestimated for Gaussians with smaller means (as compared to the other class for corresponding biomarkers) and overestimated for Gaussians with larger means. 

We refine the initial estimates using a Gaussian mixture model (GMM) and include all the available data, including MCI subjects and previously misclassified cases. To obtain a robust GMM fit, a constrained optimization method is used, with bounds on the means, standard deviations and mixing parameters, based on the aforementioned relationship between the initial estimates and their corresponding expected unbiased estimates. The objective function for optimization for biomarker $i$ is a summation of log-likelihoods, for all subjects:

\begin{equation}
\label{eq:GMMopt}
 C_i = \sum_{\forall j} \log f(x_{j,i})
\end{equation}

\noindent where the likelihood function $f(x_{j,i})$ is computed as a function of mixing parameters $(\theta^E_i, \theta^{\neg E}_i)$ for the groups corresponding to post-event and pre-event respectively and their corresponding Gaussian distributions $(\mu^E_i, \sigma^E_i)$  and $(\mu^{\neg E}_i,\sigma^{\neg E}_i)$:

\begin{multline}
\label{eq:GMMopt}
f(x_{j,i}) = \theta^E_i p(x_{j,i} | \mu^E_i, \sigma^E_i) + \theta^{\neg E}_i p(x_{j,i} | \mu^{\neg E}_i, \sigma^{\neg E}_i) \\ = \theta^E_i p(x_{j,i} | E_i) + \theta^{\neg E}_i p(x_{j,i} | \neg E_i) 
\end{multline}

\noindent $\theta^E_i$ and $\theta^{\neg E}_i$ are selected such that $\theta^E_i + \theta^{\neg E}_i = 1$. The mixing parameters and the Gaussian parameters are optimized alternately, until convergence of the mixing parameters. The initialization and optimization strategy in GMM is illustrated in Figure~\ref{fig:GMM_Illu}.

The strategy of alternating between optimizing for mixing parameter and optimizing for Gaussian parameters in combination with the initialization strategy and the subsequent constraints is different from all previous versions of EBM and it will be shown in Section~\ref{sec:results} that this results in more accurate central ordering of events in most cases.

\subsection{Estimating a disease progression timeline} \label{ssec:CentralOrdering}

\subsubsection{Estimating Subject-Specific Orderings}

The PDF thus obtained is used for classification of the biomarkers using a Bayesian classifier, where the mixing parameters ($\theta^E_i$ and $\theta^{\neg E}_i$) are used as the prior probabilities ($p(E_i)$ and $p(\neg E_i)$ respectively) when estimating posterior probabilities for each biomarker. We assume these posterior probabilities to be a measure of progression of a biomarker. Thus,  $s_j$ is established such that:

\begin{multline}
\label{eq:s_j}
s_j \ni p(E_{s_j(1)} | x_{j,s_j(1)}) > p(E_{s_j(2)} | x_{j,s_j(2)}) > \\ ... > p(E_{s_j(N)} | x_{j,s_j(N)})
\end{multline}

However, the posterior probability is influenced not only by progression of the biomarker value to its abnormal state, but also by inherent variability in normal and abnormal biomarker values across subjects, and by measurement noise. Disentangling measurement noise and inherent variability in normal biomarker values from progression of the biomarker to its abnormal state can only be done based on longitudinal data. This makes $s_j$ a noisy estimate.

\subsubsection{Estimating a central ordering}
Since the event ordering for each subject is estimated independently, any heterogeneity in disease progression is captured in the estimates of $s_j$. The central event ordering $(S)$ is the mean of the subject-specific estimates of $s_j$. To describe the distribution of $s_j$, we make use of a generalized Mallows model. The generalized Mallows model is parameterized by a central (`mean') ordering as well as spread parameters (analogous to the standard deviation in a normal distribution). The central ordering is defined as the ordering that minimizes the sum of distances to all subject-wise orderings $s_j$. To measure distance between orderings, an often used measure is Kendall's Tau distance \citep{Huang:2012}. Kendall's Tau distance between a subject specific event ordering $(s_j)$ and central ordering $(S)$ can be defined as:

\begin{equation}
\label{eq:kendall21}
K(S,s_j) = \sum_{i=1}^{N-1} V_{i}(S,s_j)
\end{equation}

\noindent where $V_{i}(S,s_j)$ is the number of adjacent swaps needed so that event at position $i$ is the same in $s_j$ and $S$. 

Since the estimates of $s_j$ are based on rankings of posterior probabilities, it would be desirable to penalize certain swaps more than others, based on how close the posterior probabilities are to each other. To this end, we introduce a probabilistic Kendall's Tau distance, which penalizes each swap based on the difference in posterior probabilities of the corresponding events. 

\begin{equation}
\label{eq:kendall22}
\widehat{K}(S,s_j) = \sum_{i=1}^{N-1} \widehat{V}_{i}(S,s_j)
\end{equation}

$\widehat{V}_{i} \forall i\in [1, N-1]$ is computed sequentially using the following algorithm\footnote{The summation symbol in step $4$ was missed accidentally in~\cite{Vikram:2017}.}:

\begin{algorithm}[H]
\caption{Probabilistic Kendall Tau distance between Subject-specific event orderings and central event ordering}\label{alg:kendtau}
\begin{algorithmic}[1]
\For{$i \in \left[1,N-1\right]$}
	\State $k \leftarrow s^{-1}_j\left(S(i)\right)$
	\If{$ k > i$}
		\State $\widehat{V}_{i}(S,s_j) \leftarrow \sum_{l = i +1}^k p_i - p_l$
		\State Move $s_j(k)$ to position $i$ and update $s_j$
	\Else
		\State $\widehat{V}_{i}(S,s_j) \leftarrow 0$
	\EndIf
\EndFor
\end{algorithmic}
\end{algorithm}

\noindent where $p_a$ is shortened notation for $p \left(E_{s_j(a)} | x_{j,s_j(a)} \right)$.

This variant of Kendall's Tau distance is quite close to the weighted Kendall's Tau distance defined in the permutation space introduced in~\cite{Kumar:2010}. The difference stems from the fact that since the probabilistic Kendall's Tau distance is between individual estimates and a central-ordering, the penalization of each swap is weighted asymmetrically as $\widehat{V}_{i}(S,s_j) \neq \widehat{V}_{i}(s_j,S)$.

The optimum $S$ is the one that minimizes $\sum_{\forall j}\widehat{K}(S,s_j)$. However, computing a global optimum $S$ based on subject-wise orderings is NP-hard. Thus getting a good initial estimate of $S$ is important to ensure the estimated $S$ is not a suboptimal local optimum. In our implementation the initial estimate of $S$ is based on ordering $\theta^{\neg E}_i$. The motivation for this is discussed in Section~\ref{ssec:PatientStaging}. $S$ was further optimized based on the algorithm introduced by ~\cite{Fligner:1988} to estimate the central ordering.

\subsubsection{Estimating Event Centers}

The $S$ that has been derived in this manner, is an estimate of the sequence in which the biomarkers become abnormal during the progression of a disease. However, it falls short of being a disease timeline, because it does not provide information about the proximity of consecutive events. To address this issue, we estimate distances between events by computing the cost of adjacent swaps in the event ordering, as measured by summation of probabilistic Kendall's Tau distance over all subjects.

\begin{equation}
\label{eq:swap}
\Gamma_{i+1,i} = \sum_{\forall j} \widehat{K}(S_{i+1,i},s_j) - \widehat{K}(S,s_j)
\end{equation}

\noindent where $S_{i+1,i}$ is identical to $S$ except for the swap between events at locations $i$ and $i+1$, and $\Gamma_{i+1,i}$ is the cost of the swap. This represents the cost for the central ordering to be $S_{i+1,i}$ instead of $S$. We hypothesize that the closer the events $i+1$ and $i$ are to each other, the lower the swapping cost would be. Hence we consider these costs to be proportional to distance between events in terms of biomarker progression.

To estimate the distance of the first biomarker being abnormal (event) in $S$ to a hypothetical disease-free individual, we introduce a pseudo-event which becomes abnormal at the beginning of the disease timeline and hence is abnormal for all the subjects in the database \textit{i.e.} $p \left(E_{0} | x_{j,0} \right) = 1$  $\forall j$. Similarly, we introduce another pseudo-event which becomes abnormal at the end of the disease timeline and hence is normal for all the subjects in the database \textit{i.e.} $p \left(E_{N+1} | x_{j,N+1} \right) = 0$ $\forall j$. We scale $\Gamma_{i+1,i} \forall i \in [0,N]$ such that $\sum \overline{\Gamma}_{i+1,i} = 1$. Event center $(\lambda_k)$ of event $k$ in $S$ for $k > 0$, is computed as follows:

\begin{equation}
\label{eq:ec}
\lambda_k = \sum_{i=0}^{k-1} \overline{\Gamma}_{i+1,i}
\end{equation}

In fact, the concept of event centers can also be extended to Fonteijn's EBM by computing the cost of adjacent swaps in the event ordering as the difference in log-likelihoods as follows:

\begin{equation}
\label{eq:swap2}
\Gamma_{i+1,i} = \log\left(p(X|S)\right) - \log\left(p(X|S_{i+1,i})\right) 
\end{equation}

\noindent Extension of this concept to Huang's EBM is not straightforward and is beyond this paper's scope.

The set of event centers $\lambda_{1,2,...,N}$, will henceforth be referred to as $\Lambda$. This results in a disease timeline, with $S$ giving information about the order of progression of biomarkers and $\Lambda$ giving information about the event centers in this timeline.

\subsection{Patient Staging} \label{ssec:PatientStaging}

Once the central ordering of events $(S)$ and event centers $(\Lambda)$ have been determined, we propose a patient staging algorithm where a patient stage $(\Upsilon_j)$ is interpreted as an expectation of $\lambda_k$ with respect to the conditional distribution $p(k|S,X_j)$. Thus, $\Upsilon_j$ can be written as given below:

\begin{equation}
\label{eq:pstage12}
\Upsilon_j = \frac{ \sum_{k=1}^N \lambda_k  p(k | S,X_j)}{\sum_{k=1}^N p(k | S,X_j)}
\end{equation}

\noindent Multiplying $p(S,X_j)$ in both numerator and denominator and using the chain rule of probability results in:

\begin{equation}
\label{eq:pstage13}
\Upsilon_j = \frac{ \sum_{k=1}^N \lambda_k  p(k,S,X_j)}{\sum_{k=1}^N p(k,S,X_j)}
\end{equation}

\noindent Using chain rule of probability, we can write $p(k,S,X_j)$ as:

\begin{equation}
\label{eq:pstage}
p(k,S,X_j) = p(X_j|k,S)  p(k,S)
\end{equation}

If we assume a uniform distribution of $p(k|S)$ and $p(S)$ as in~\cite{Fonteijn:2012}, $p(k,S,X_j)$ becomes equal to $p(X_j|k,S)$, which was used for patient staging in Fonteijn's EBM as discussed in Section~\ref{sec:relatedwork}. However we use prior knowledge in order to define a more informative distribution $p(k,S)$:

\begin{equation}
\label{eq:psigmak}
p(k , S) = \frac{\prod_{i=1}^k \theta^E_{S(i)} \prod_{i=k+1}^N \theta^{\neg E}_{S(i)}}{Z}
\end{equation}

\noindent where $Z$ is a normalizing factor, chosen so as to make this a probability. This choice of $p(k , S)$ can be justified because biomarkers which become abnormal earlier in the disease process are more likely to have a higher value of $\theta^E_i$ than the biomarkers which become abnormal later. Hence it is far more likely to have a central-ordering based on ascending values of $\theta^{\neg E}_i$ than an ordering with ascending values of $\theta^E_i$. It should be noted that, the choice of $p(k , S)$ is not unique. For example, it could also be any $n$-th power of the above equation $\forall n > 0$. Thus, from Equations~\ref{eq:pstage}, ~\ref{eq:psigmak} and ~\ref{eq:font12}, we get:

\begin{multline}
\label{eq:pstage3}
p\left(k, S , X_j \right) \propto \prod_{i=1}^k p \left( x_{j,S(i)} | E_{S(i)} \right) \theta^E_{S(i)} \times \\ \prod_{i=k+1}^N p \left( x_{j,S(i)} | \neg E_{S(i)} \right) \theta^{\neg E}_{S(i)}
\end{multline}

Using the above value of $p\left(k, S , X_j \right)$ in Equation~\ref{eq:pstage13}, results in continuous patient stages.

\section{Experiments}  \label{sec:exp}

This section describes the experiments performed to benchmark the accuracy of the proposed DEBM algorithm and compare it with state-of-the-art EBM methods. The EBM methods used for comparison in these experiments are Huang's EBM~\citep{Huang:2012} and the variant of Fonteijn's EBM that is suited for AD disease progression modeling~\citep{Young:2014}. The source code for DEBM and Fonteijn's EBM, with different mixture modeling techniques and patient staging techniques discussed in this paper have been made publicly available online under the GPL 3.0 license: \url{https://github.com/88vikram/pyebm/}. The source code for Huang's EBM used in our experiments was provided by the authors of the method.

For brevity, Fonteijn's EBM and Huang's EBM will henceforth be referred to as FEBM and HEBM, respectively. The mixture model used with an EBM model (as the one described in Section ~\ref{ssec:subjectordering}) will be denoted by a subscript. For example, FEBM with the Gaussian mixture model proposed in~\cite{Young:2014} will be referred to as FEBM\textsubscript{ay}. The Gaussian mixture model optimization techniques in~\cite{Huang:2012}, ~\cite{Vikram:2017} and the one introduced in this paper will be denoted with subscripts `jh', `vv1' and `vv2' respectively.\footnote{Mixture model `ay' optimizes for Gaussian and mixing parameters together. Initialization of Gaussian parameters for optimization is done without rejecting the overlapping part of Gaussians in CN and AD classes. `vv1' also optimizes for Gaussian and mixing parameters together (although with much stricter bounds) but the initialization of Gaussian parameters is similar to the one in this paper. `jh' couples mixture modeling with estimation of subject-specific ordering to estimate a combined optimum solution.}

Data used in the experiments were obtained from the Alzheimer’s Disease Neuroimaging Initiative (ADNI) database (adni.loni.usc.edu) \footnote{The ADNI was launched in 2003 as a public-private partnership, led by Principal Investigator Michael W. Weiner, MD. The primary goal of ADNI has been to test whether serial magnetic resonance imaging (MRI), positron emission tomography (PET), other biological markers, and clinical and neuropsychological assessment can be combined to measure the progression of mild cognitive impairment (MCI) and early Alzheimer’s disease (AD). For up-to-date information, see www.adni-info.org.} . We begin with the details of the experiments performed on ADNI data to estimate the event ordering in Section~\ref{ssec:adniexp}. Since the ground-truth event ordering is unknown for clinical datasets, we resort to using the ability of patient staging to classify AD and CN subjects, as an indirect way of measuring the reliability of the event ordering. We also measure the accuracy of event ordering and relative distance between events more directly by performing extensive experiments on synthetic data simulating the progression of AD. The details of these experiments are given in Section~\ref{ssec:simulexp}.

\subsection{ADNI Data} \label{ssec:adniexp}

We considered 1737 ADNI subjects (417 CN, 978 MCI and 342 AD subjects) who had a structural MRI (T1w) scan at baseline. The T1w scans were non-uniformity corrected using the N3 algorithm~\citep{Tustison:2010}. This was followed by multi-atlas brain extraction using the method described in~\cite{Bron:2014}. Multi-atlas segmentation was performed~\citep{Hammers:2003,Gousias:2008} using the structural MRI scans to obtain a region-labeling for $83$ brain regions in each subject using a set of $30$ atlases. Probabilistic tissue segmentations were obtained for white matter, gray matter (GM), and cerebrospinal fluid on the T1w image using the unified tissue segmentation method~\citep{Ashburner:2005} of SPM8 (Statistical Parametric Mapping, London, UK). The probabilistic GM segmentation was then combined with region labeling to obtain GM volumes in the extracted regions. We also downloaded CSF (A$\beta_{1-42}$ (ABETA), TAU and p-TAU) and cognitive score (MMSE, ADAS-Cog) values from the ADNI database, making the total number of features equal to $88$. 

The features TAU and p-TAU were transformed to logarithmic scales to make the distributions less skewed. GM volumes of segmented regions were regressed with age, sex and intra-cranial volume (ICV) and the effects of these factors were subsequently corrected for, before being used as biomarkers. The effect of age and sex was regressed out of CSF based features, whereas effects of age, sex and education was regressed out of cognitive scores. 

We retained $52$ biomarkers (GM volume based biomarkers of $47$ regions, $3$ CSF and $2$ cognitive scores) having significant differences between CN and AD subjects using Student's t-test with $p<0.005$, after Bonferroni correction. These biomarker values were used to perform two sets of experiments.

\textbf{Experiment 1(a):} A subset of $7$ biomarkers including the $3$ CSF features, MMSE score, ADAS-Cog score, gray matter volume of the hippocampus (combined volume of left and right hippocampi) and gray matter volume in whole brain was created. Event ordering of these $7$ biomarkers was inferred using DEBM. We studied the positional variance of central ordering and variance of event centers inferred by DEBM by creating $100$ bootstrapped samples of the data. 

\textbf{Experiment 1(b):} The Biomarkers were ranked based on their aforementioned $p$-value and the above experiment was repeated with top $25$ and top $50$ biomarkers to investigate if the event-centers estimated for the subset of Biomarkers used in Experiment 1(a), remain comparable to the ones estimated in Experiment 1(a). 

\textbf{Experiment 2:} As an indirect way of measuring the accuracy of the estimated event ordering, we use patient staging based on the estimated event orderings as a way to classify CN and AD subjects in the database. $10$-fold cross validation was used for this purpose. AUC measures were used to measure the performance of these classifications and thus indirectly hint at the reliability of the event ordering based on which the corresponding patient staging were performed. We used varying number of biomarkers (ranked based on their p-value) ranging from $5$ to $50$ in steps of $5$ for this experiment. We used the methods FEBM\textsubscript{ay}, HEBM\textsubscript{jh}, DEBM\textsubscript{vv1} and DEBM\textsubscript{vv2} for inferring the ordering. Patient staging was done based on the methods described in their respective papers. Since the earlier version of DEBM~\citep{Vikram:2017} had not introduced a patient staging method, we use the patient staging method described in this paper for evaluating the method.

\subsection{Simulation Data} \label{ssec:simulexp}

We used the framework developed by~\cite{Young:2015} for simulating cross-sectional data consisting of scalar biomarker values for CN, MCI and AD subjects. In this framework, disease progression in a subject is modeled by a cascade of biomarkers becoming abnormal and individual biomarker trajectories are represented by a sigmoid. The equation for generating biomarker values for different subjects is given below:

\begin{equation}
\label{eq:sigmoid}
x_{j,i}(\Psi) = \frac{R_{i}}{1+ \exp(-\rho_{i}(\Psi - \xi_{j,i}))} + \beta_{j,i}
\end{equation}

\noindent $\Psi$ denotes disease stage of a subject which we take to be a random variable distributed uniformly throughout the disease timeline. $\rho_{i}$ signifies the rate of progression of a biomarker, which we take to be equal for all subjects. $\xi_{j,i}$ denotes the disease stage at which the biomarker becomes abnormal. $\beta_{j,i}$ denotes the value of the biomarker when the subject is normal and $R_{i}$ denotes the range of the sigmoidal trajectory of the biomarker, which we take to be equal for all subjects. 

In our experiments, $\beta_{j,i}$ and $\xi_{j,i}$ $\forall j$ are assumed to be random variables with Normal distribution $\mathbb{N}(\mu_{\beta_i},\Sigma_{\beta_i})$ and $\mathbb{N}(\mu_{\xi_i},\Sigma_{\xi_i})$ respectively. $\mu_{\beta_i}$ is equal to the mean value of the corresponding biomarker in the CN group of the selected ADNI data. $R_{i}$ is equal to the difference between the mean values of the biomarker in the CN and AD groups of the selected ADNI data. $\Sigma_{\beta_i}$ represents the variability of biomarker values in the CN group. We consider a relative scale for $\Sigma_{\beta_i}$, where $1$ refers to the observed variation among the CN subjects in ADNI data. Variation in $\xi_{j,i}$ is controlled by $\Sigma_{\xi_i}$ and results in variation in ordering among subjects in population and could be seen as a parameter controlling the disease heterogeneity within a simulated population. $\Sigma_{\xi_i}$ $\forall i$ is varied in multiples of $\Delta \xi$, where $\Delta \xi$ is the average difference between adjacent $\mu_{\xi_i}$. $\mu_{\xi_i}$ refers to the event centers of various biomarkers. The set of $\mu_{\xi_i} \forall i$ will collectively be referred to as $\Lambda_{gt}$ and they will be used to assess the accuracy of estimated event centers $(\lambda_i)$.

The parameters in the simulation framework that could have an effect on the performance of EBMs are $\Sigma_{\beta_i}$, $\mu_{\xi_i}$, $\Sigma_{\xi_i}$, and $\rho_{i}$. Apart from this, the number of subjects $(M)$ and the number of biomarkers $(N)$ in the dataset could also have an effect on the performance of EBMs. Using this simulation framework, we study the effect of the aforementioned parameters on the ability of different variants of EBM algorithms to accurately infer the ground-truth central ordering in the population. Change in $\mu_{\beta_i}$ results only in a translational effect on biomarker values and change in $R_i$ results only in a scaling effect on biomarker values. These factors do not affect the performance of the EBMs and hence were not evaluated in our experiments.

Performance of an EBM method can be measured using error in estimation of either $S$ or $\Lambda$. Error in estimating $S$ $(\epsilon_{S})$ will henceforth be referred to as `ordering error' whereas the error in estimating $\Lambda$ $(\epsilon_{\Lambda})$ will henceforth be referred to as `event-center error'. $\epsilon_{S}$  is computed using the following equation:

\begin{equation}
\label{eq:E1}
\epsilon_{S} = \frac{K(S,S_{gt})}{\binom{N}{2}}
\end{equation}

\noindent where $S_{gt}$ is the ground truth ordering. $\epsilon_{S}$ is effectively a normalized Kendall's Tau distance between $S$ and $S_{gt}$. The normalization factor for $\binom{N}{2}$, was chosen to make the accuracy measure interpretable for different number of biomarkers. 

For comparing $\Lambda$ and $\Lambda_{gt}$, $\Lambda$ were scaled and translated such that the mean and standard deviation of $\Lambda$ were equal to that of $\Lambda_{gt}$. This is done because we are only interested in evaluating the errors in estimating relative distance between events and not the absolute position of event-centers. The choice of scale in event-centers are arbitrary and the chosen scale for the estimated event-centers was based on pseudo-events, which need not necessarily coincide with the simulation framework's ground-truth event-centers.

\begin{equation}
\label{eq:E2}
\epsilon_{\Lambda} = \sum_{\forall i} | \lambda_{i}^{st} - \mu_{\xi_{i}} |
\end{equation}

\noindent where $\lambda_{i}^{st}$ is the scaled and translated version of $\lambda_{i}$.

As mentioned before, the factors that can have an effect on the performance of EBMs are $\Sigma_{\beta_i}$, $\mu_{\xi_i}$, $\Sigma_{\xi_i}$, $\rho_{i}$, $M$ and $N$. In each of the following $5$ experiments, a few of these factors were varied while the others were set to their default values. The default value for $\Sigma_{\beta_i}$ was taken to be $1$ as this corresponds to the observed variation among CN subjects in ADNI. $\mu_{\xi_i}$ were spaced equidistantly, i.e., $\mu_{\xi_{i+1}} - \mu_{\xi_i} = 1/(N+1)$. As the actual variation in event centers among different subjects is not known in a clinical dataset, the default value of $\Sigma_{\xi_i}$ was taken to be $2 \Delta \xi$. For the sake of simplicity of notation $\Delta \xi$ will be omitted henceforth, and the values of $\Sigma_{\xi_i}$ are implicitly in multiples of $\Delta \xi$. $\rho_{i}$ was considered to be equal for all biomarkers by default. The default values for $M$ and $N$ were $1737$ and $7$ respectively, mimicking the dataset used in Experiment 1(a). For each simulation setting, $50$ repetitions of simulation data were created and used for benchmarking the performance of EBMs on synthetic data.

\textbf{Experiment 3:} The first simulation experiment was performed to study the effect of $\Sigma_\beta \in [0.2,1.8]$ and $\Sigma_\xi \in [0,4]$, varying one at a time while keeping the other at its mean value. The $\epsilon_S$ of FEBM\textsubscript{ay}, FEBM\textsubscript{vv2}, HEBM\textsubscript{jh}, HEBM\textsubscript{vv2}, DEBM\textsubscript{vv1} and DEBM\textsubscript{vv2} were determined. 

\textbf{Experiment 4:} The above experiment was repeated for DEBM\textsubscript{vv2} and FEBM\textsubscript{vv2} and the $\epsilon_\Lambda$ were measured for the two methods.

\textbf{Experiment 5:} This experiment was performed to study the effect of a non-uniform distribution of $\mu_{\xi_i}$. $\Sigma_\beta$ and $\Sigma_\xi$ combinations of $(0.6,1)$, $(1.0,2)$, $(1.4,3)$ and $(1.8,4)$ were tested to study their effect in non-uniformly spaced biomarkers. $\epsilon_S$ of DEBM\textsubscript{vv2}, FEBM\textsubscript{vv2} and HEBM\textsubscript{vv2} were measured. Additionally, $\epsilon_\Lambda$ of DEBM\textsubscript{vv2} and FEBM\textsubscript{vv2} were measured. To also study the effect of unequal rates of progression of biomarkers ($\rho_{i}$), the above experiment was performed once with equal $\rho_{i}$ for all biomarkers and once when they were unequal. The experiment with unequal biomarker rates had the same mean biomarker progression rate as the the experiment with equal biomarker rates. The progression rates of different biomarkers has been included as supplementary material (Figure S1).


\textbf{Experiment 6:} This experiment was performed to study the influence of the number of subjects $(M)$. $M$ was varied from $100$ to $2100$ in steps of $200$. $\epsilon_S$ of DEBM\textsubscript{vv2}, FEBM\textsubscript{vv2} and HEBM\textsubscript{vv2} were measured. DEBM\textsubscript{vv2} and FEBM\textsubscript{vv2} were also assessed based on $\epsilon_\Lambda$.

\textbf{Experiment 7:} This experiment was performed to study the influence of the number of biomarkers $(N)$. $N$ was varied from $7$ to $52$ in steps of $5$. In each random generation of a dataset, we randomly selected (with replacement) the biomarkers to be used in the iteration. This was done to study the effect of $N$ on the EBM models and separate it from the effect of adding weaker biomarkers. $\epsilon_S$ of DEBM\textsubscript{vv2}, FEBM\textsubscript{vv2} and HEBM\textsubscript{vv2} were measured. DEBM\textsubscript{vv2} and FEBM\textsubscript{vv2} were also assessed based on $\epsilon_\Lambda$.

\begin{figure*}[p!]
    \centering
        \includegraphics[width=0.8\textwidth]{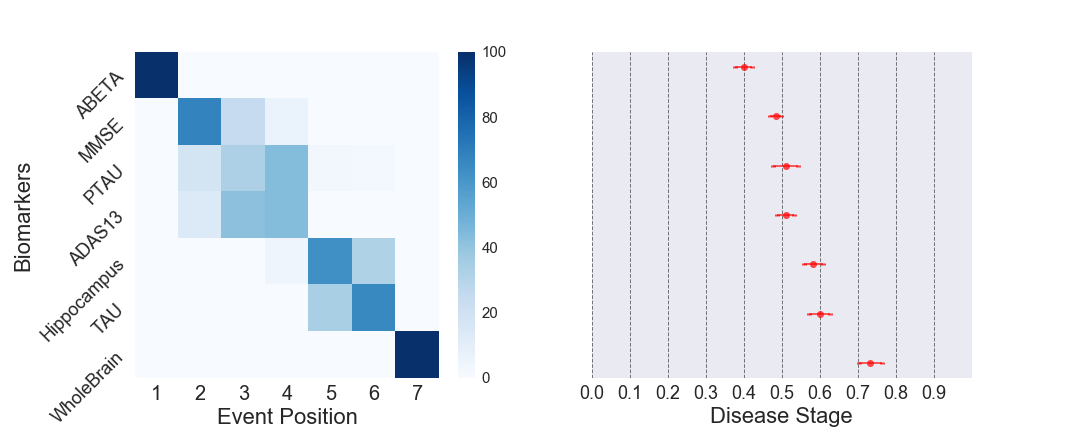}
        
        \caption{Experiment 1(a): DEBM\textsubscript{vv2} with 7 Events. The positional variance diagram (left) shows the uncertainty in estimating the central event ordering. The event-center variance diagram (right) shows the standard error of estimated event centers. These were measured by 100 repetitions of bootstrapping.}
        \label{fig:ADNI7}
\end{figure*}

\begin{figure*}[p!]
    \centering
        \includegraphics[width=0.8\textwidth]{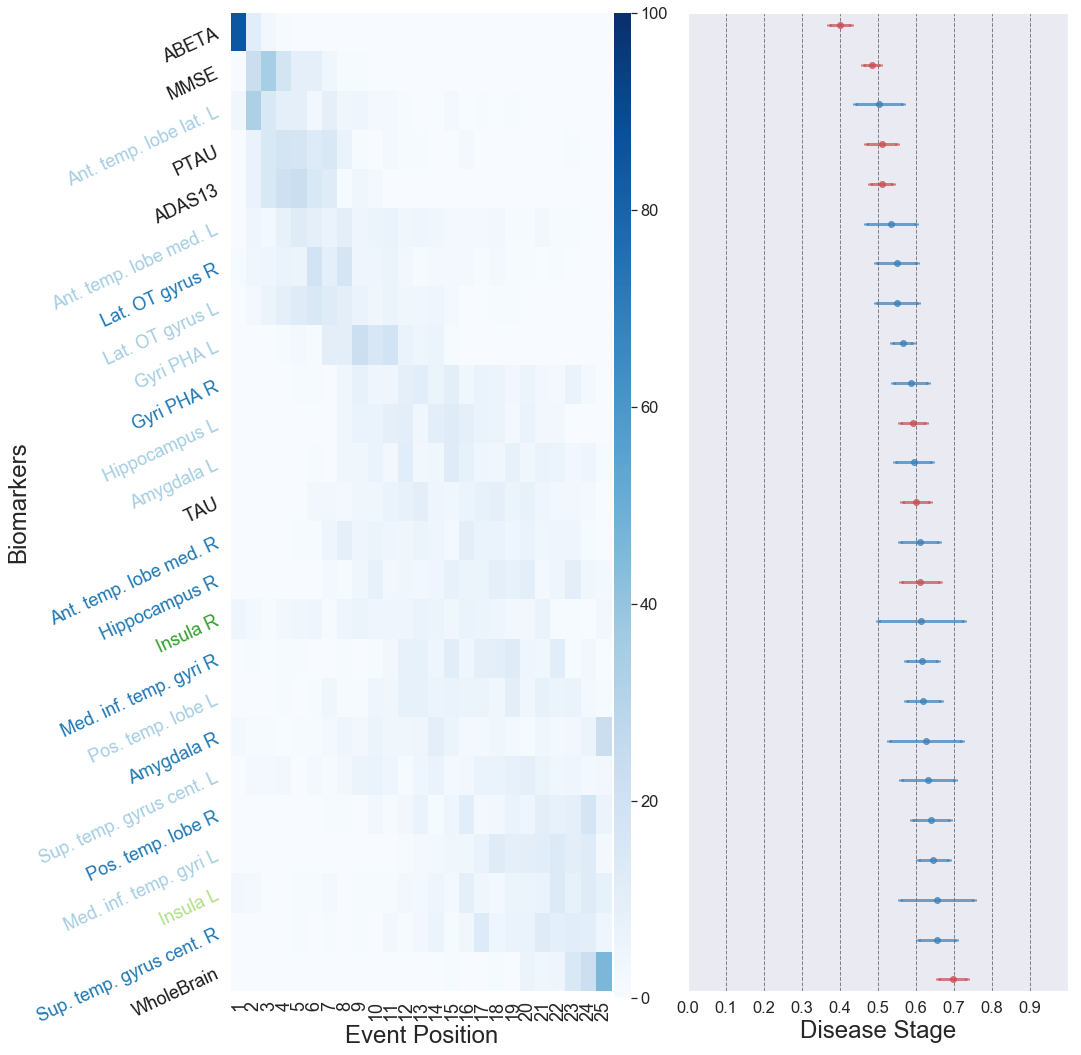}
        
        \caption{Experiment 1(b): DEBM\textsubscript{vv2} with $25$ Events. The positional variance diagram (left) shows the uncertainty in estimating the central event ordering and the event-center variance diagram (right) shows the standard error of estimated event centers. These were measured by $100$ repetitions of bootstrapping. The event centers of the biomarkers used in Figure~\ref{fig:ADNI7} are marked in red. Table~\ref{tab:Abbreviation} shows the full forms of the abbreviations used in the y-axis labels. Figure~\ref{fig:legend} maps the colors used for y-axis labels to different lobes in the brain.}
        \label{fig:ADNI25}
\end{figure*}        
        
\begin{figure*}[h!]
    \centering
        \includegraphics[width=0.85\textwidth]{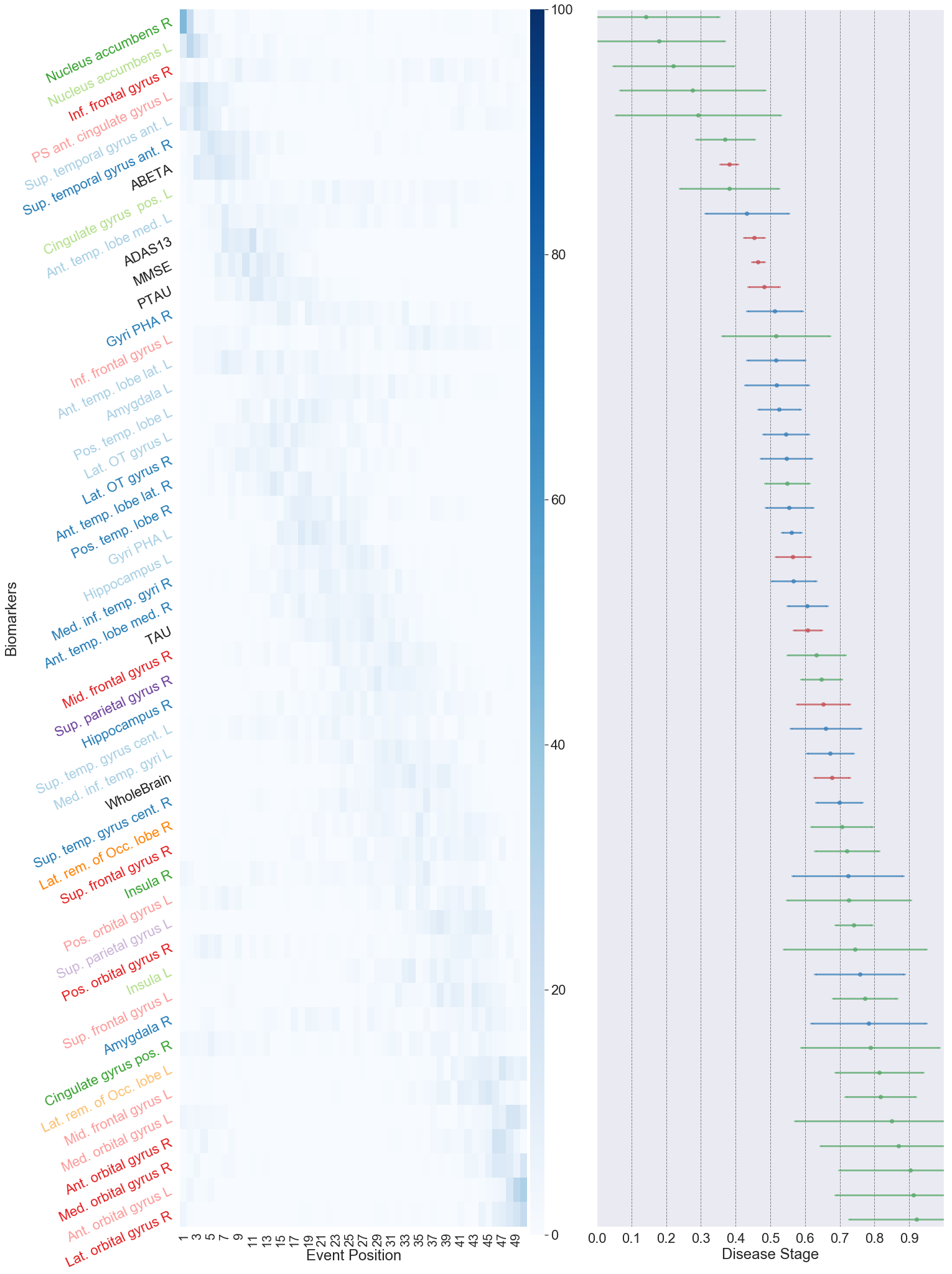}
        
        \caption{Experiment 1(b): DEBM\textsubscript{vv2} with $50$ Events. Positional variance diagram (left) shows the uncertainty in estimating the central event ordering and event center variance diagram (right) shows the standard error of estimated event-centers. These were measured by $100$ repetitions of bootstrapping. The event-centers of the biomarkers used in Figure~\ref{fig:ADNI7} are marked in red, whereas the ones used in Figure~\ref{fig:ADNI25} are marked in blue. Table~\ref{tab:Abbreviation} shows the full forms of the abbreviations used in the y-axis labels. Figure~\ref{fig:legend} maps the colors used for y-axis labels to different lobes in the brain.}
        \label{fig:ADNI50}
\end{figure*}

\section{Results} \label{sec:results}

\subsection{ADNI Data} \label{ssec:resadni}

\begin{figure}[t!]
\centering
\begin{tikzpicture}
  \node (img)  {\includegraphics[width=0.45\textwidth]{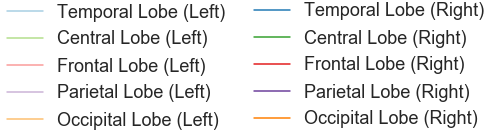}};
\end{tikzpicture}
\caption{Legend for the colors used in Figures~\ref{fig:ADNI25} and ~\ref{fig:ADNI50}. The colors map different biomarker labels to lobes in the brain.}
\label{fig:legend}
\end{figure}

\textbf{Experiment 1:} Figures~\ref{fig:ADNI7},~\ref{fig:ADNI25} and~\ref{fig:ADNI50} show the positional variance and event-center variance obtained using DEBM\textsubscript{vv2} with $7$, $25$ and $50$ number of events respectively. Table~\ref{tab:Abbreviation} shows the abbreviations used in Figures~\ref{fig:ADNI25} and~\ref{fig:ADNI50} along with their full names. Figure~\ref{fig:legend} maps the colors used for y-axis labels of Figures~\ref{fig:ADNI25} and~\ref{fig:ADNI50} to different lobes in the brain. 

It can be seen from Figure~\ref{fig:ADNI7} (left) that CSF-based biomarkers ABETA becomes abnormal before MMSE and CSF-based p-TAU. This is followed by ADAS13, Hippocampal volume, TAU and whole brain volume. However Figure~\ref{fig:ADNI7} (right) shows that the event centers for MMSE, ADAS13, p-TAU are close to each other and so are the event-centers of TAU and hippocampus volume. The event associated with the TAU biomarker seems closer to the whole brain volume event as they are in positions $6$ and $7$ of Figure~\ref{fig:ADNI7} (left). However, the centers of these two events are quite far apart in Figure~\ref{fig:ADNI7} (right) and the p-TAU event (position 2) is closer to the TAU event than whole brain volume event.

\begin {table}
\footnotesize
\begin{center}
 \begin{tabular}{||c c||} 
 \hline
 Abbreviation & Full name \\ [0.5ex] 
 \hline\hline
 L & Left \\
 R & Right \\
 PHA & Parahippocampalis et Ambiens \\
 Med. & Medial \\ 
 Inf. & Inferior \\
 Sup. & Superior \\
 Temp. & Temporal \\ 
 Pos. & Posterior \\
 Lat. & Lateral \\
 Ant. & Anterior \\
 OT & Occipitotemporal \\
 Cent. & Central \\
 Mid. & Middle \\
 Rem. & Remainder \\
 Occ. & Occipital \\
 PS & Pre-subgenual \\
 \hline
 
\end{tabular}
\end{center}
\caption{Abbreviations used in Figures~\ref{fig:ADNI25} and ~\ref{fig:ADNI50} along with their full names~\citep{Hammers:2003}.}
\label{tab:Abbreviation} 
\end{table}

As the number of biomarkers increases, the variation in the positions also increases considerably, as seen in Figures~\ref{fig:ADNI25} (left) and ~\ref{fig:ADNI50} (left). The event centers of the biomarkers used in Experiment 1(a) remain fairly consistent ($\pm 0.05$) in Experiment 1(b). It can also be seen that biomarkers with lower $p$-values (biomarkers included in the model with $50$ biomarkers and not in the model with $25$ biomarkers), have larger variance in their event-center estimation.

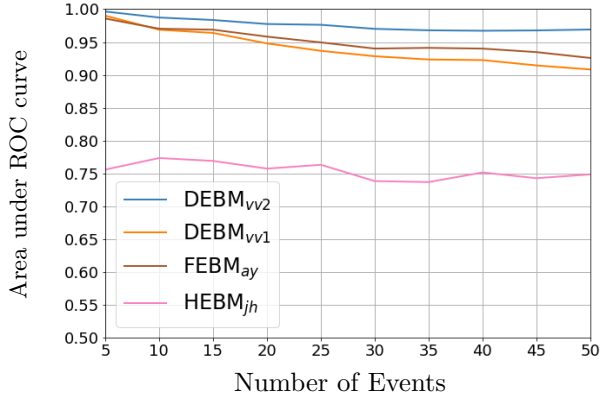
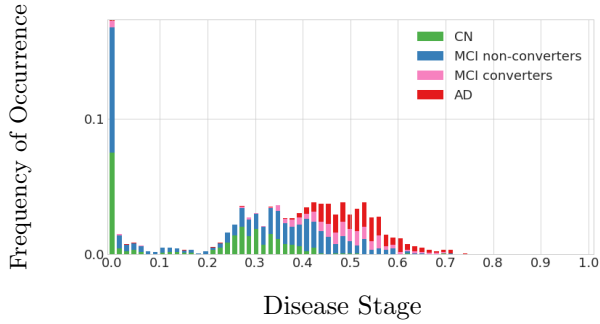
\begin{figure}[t!]
    \begin{subfigure}[b]{0.5\textwidth}
    \centering
\begin{tikzpicture}
  \node (img)  {\includegraphics[width=1\textwidth]{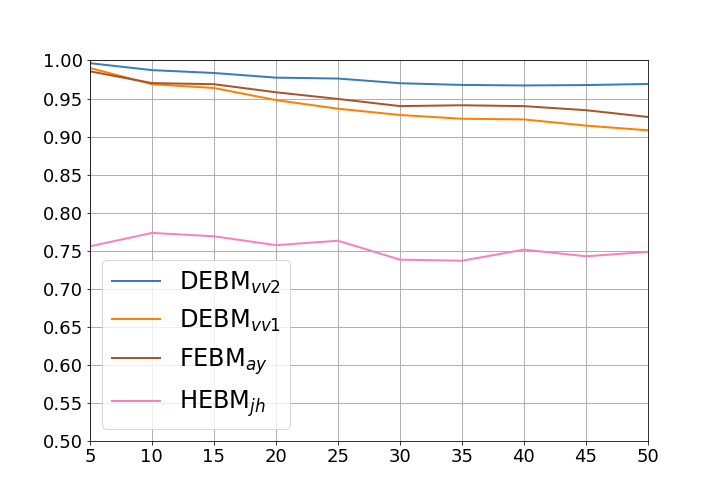}};
  \node[below=of img, node distance=0cm, yshift=1.5cm,font=\color{black}] {Number of Events};
  \node[left=of img, node distance=0cm, rotate=90, anchor=center,yshift=-1cm,font=\color{black}] {\small Area under ROC curve};
 \end{tikzpicture}
   \caption{}
\end{subfigure}

    \begin{subfigure}[b]{0.5\textwidth}
    \centering
\begin{tikzpicture}
  \node (img)  {\includegraphics[width=1\textwidth]{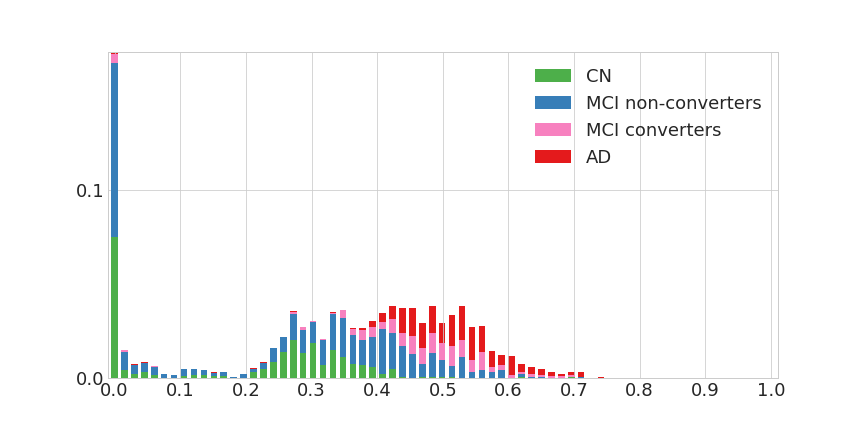}};
  \node[below=of img, node distance=0cm, yshift=1.2cm,font=\color{black}] {Disease Stage};
  \node[left=of img, node distance=0cm, rotate=90, anchor=center,yshift=-1cm,font=\color{black}] {\small Frequency of Occurrence};
\end{tikzpicture}
  \caption{}
\end{subfigure}

\caption{ Experiment 2: In (a) we see the variation of AUC with respect the number of biomarkers used for building the model using DEBM, when the obtained patient stages were used for classification of CN versus AD subjects. The AUC measure was obtained using $10$-fold cross-validation. In (b) we see the frequency of occurrence of subjects in different disease stages, when the most significant $25$ features were given as input to DEBM\textsubscript{vv2} for inferring the ordering as well as for patient staging.}
\label{fig:ADNICrossval}
\end{figure}

\textbf{Experiment 2:} Figure~\ref{fig:ADNICrossval} (a) shows the mean AUC when using patient stages for classifying CN versus AD subjects using DEBM and other variants of EBM methods. It can be observed that the AUC of all the methods decreases as the number of events increases. The proposed method DEBM\textsubscript{vv2} followed by the proposed patient staging algorithm outperforms all the existing EBM variants consistently. 

Figure~\ref{fig:ADNICrossval} (b) shows the distribution of patient stages for the whole population when the most significant $25$ features were given as input to DEBM\textsubscript{vv2}. This graph shows a peak at disease stage $0$ dominated by CN and MCI non-converters, which shows that these subjects are not progressing towards AD. The non-zero lower disease stages are dominated by CN subjects and MCI non-converters, whereas MCI converters\footnote{MCI converters are subjects who convert to AD within 3 years of baseline measurement} and the subjects with AD have higher disease stages. 

\subsection{Simulation Data} \label{ssec:resadni}

\begin{figure}[t!]
    \begin{subfigure}[b]{0.5\textwidth}
    \centering
\begin{tikzpicture}
  \node (img)  {\includegraphics[width=1\textwidth]{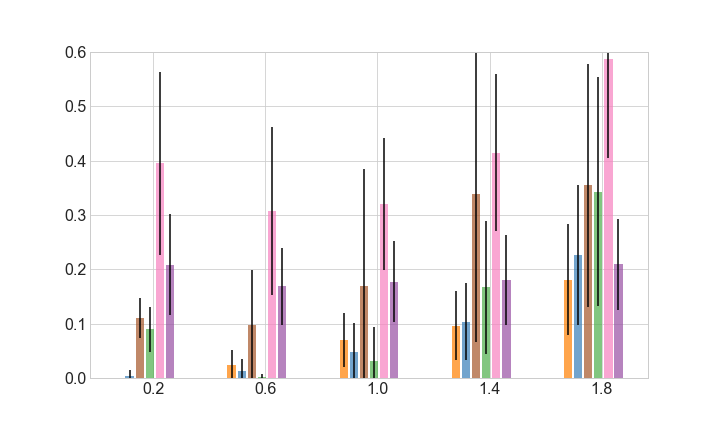}};
  \node[below=of img, node distance=0cm, yshift=1.5cm,font=\color{black}] {$\Sigma_\beta$};
  \node[left=of img, node distance=0cm, rotate=90, anchor=center,yshift=-1.5cm,font=\color{black}] {\small Ordering Error};
 \end{tikzpicture}
  \caption{}
\end{subfigure}

\begin{subfigure}[b]{0.5\textwidth}
    \centering
\begin{tikzpicture}
  \node (img)  {\includegraphics[width=1\textwidth]{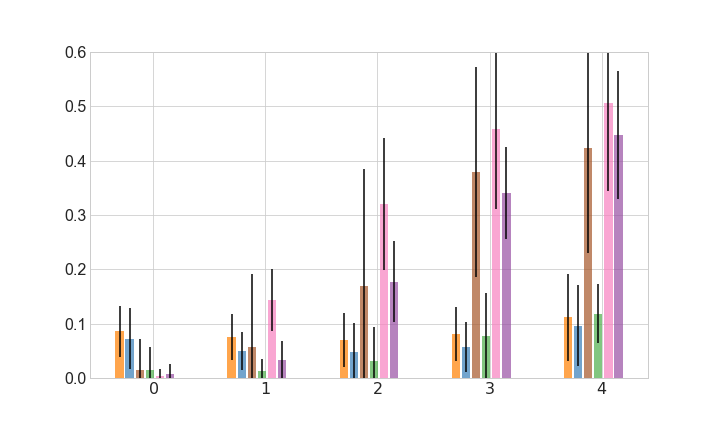}};
  \node[below=of img, node distance=0cm, yshift=1.5cm,font=\color{black}] {$\Sigma_\xi$};
  \node[left=of img, node distance=0cm, rotate=90, anchor=center,yshift=-1.5cm,font=\color{black}] {\small Ordering Error};
\end{tikzpicture}
 \caption{}
\end{subfigure}

\begin{subfigure}[b]{0.5\textwidth}
    \centering
\begin{tikzpicture}
  \node (img)  {\includegraphics[width=0.9\textwidth]{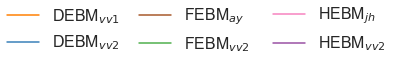}};
\end{tikzpicture}
 \caption{}
\end{subfigure}

\caption{Experiment 3: Ordering errors of DEBM\textsubscript{vv1}, DEBM\textsubscript{vv2}, FEBM\textsubscript{ay}, FEBM\textsubscript{vv2}, HEBM\textsubscript{jh} and HEBM\textsubscript{vv2} for 50 repetitions of simulations. Figure (a) shows the ordering error as a function of variability in population ($\Sigma_\beta$). Figure (a) shows the ordering error as a function of variation in ordering ($\Sigma_\xi$). Error bars in (a) and (b) represent standard deviations over the $50$ repetitions. Figure (c) shows the legend for the plots in (a) and (b).}
\label{fig:VaryNoise1}
\end{figure}
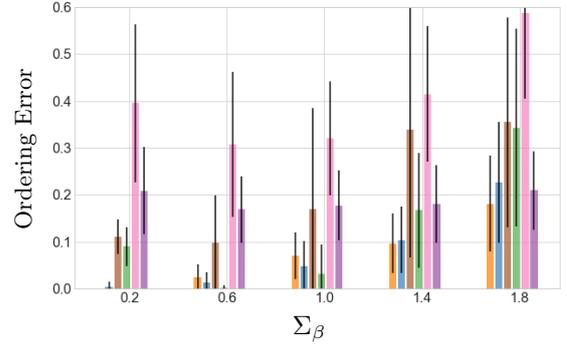
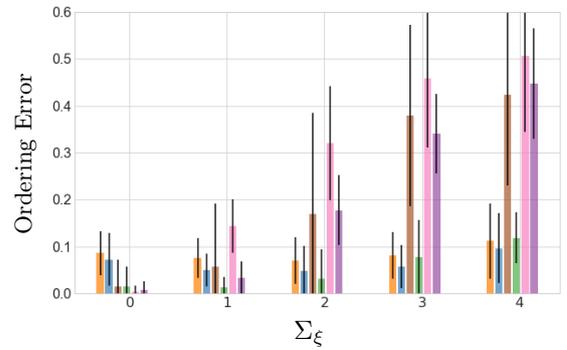
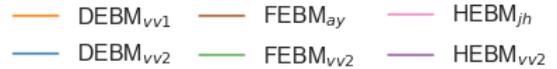

\textbf{Experiment 3:} Figures~\ref{fig:VaryNoise1} shows the ordering errors of DEBM, FEBM and HEBM models with different mixture models as $\Sigma_\beta$ and $\Sigma_\xi$ increase. The error-bars depict mean and standard deviation of the errors obtained in $50$ repetitions of simulations. It can be seen that the proposed optimization technique improves the performance of all three EBM models. The change is particularly evident when comparing the performance of FEBM\textsubscript{vv2} and FEBM\textsubscript{ay}. 

It can also be seen that FEBM\textsubscript{vv2} performs slightly better than DEBM\textsubscript{vv2} when $\Sigma_\xi$ is low, but as $\Sigma_\xi$ increases, the performance of FEBM\textsubscript{vv2} degrades significantly. The performance of HEBM is almost always worse than its FEBM or DEBM counterpart.

\begin{figure}[t!]
    \centering
    \begin{subfigure}[b]{0.45\textwidth}
    \centering
\begin{tikzpicture}
  \node (img)  {\includegraphics[width=0.85\textwidth]{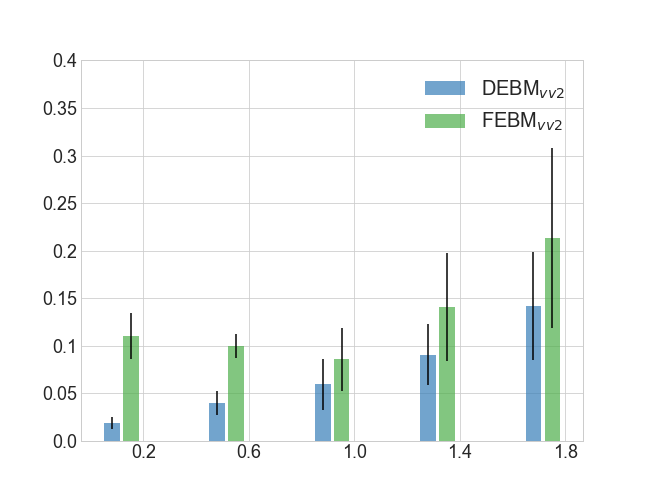}};
  \node[below=of img, node distance=0cm, yshift=1.7cm,font=\color{black}] {$\Sigma_\beta$};
  \node[left=of img, node distance=0cm, rotate=90, anchor=center,yshift=-1cm,font=\color{black}] {\small Event-Center Error};
 \end{tikzpicture}
 \caption{}
\end{subfigure}

\begin{subfigure}[b]{0.45\textwidth}
\centering
\begin{tikzpicture}
  \node (img)  {\includegraphics[width=0.85\textwidth]{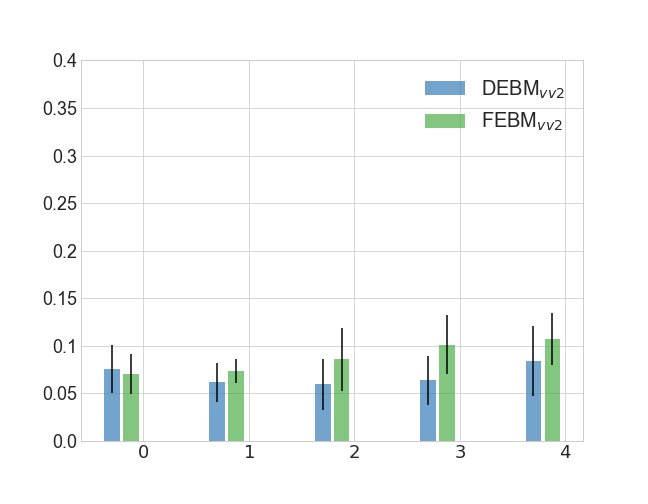}};
  \node[below=of img, node distance=0cm, yshift=1.7cm,font=\color{black}] {$\Sigma_\xi$};
  \node[left=of img, node distance=0cm, rotate=90, anchor=center,yshift=-1cm,font=\color{black}] {\small Event-Center Error};
\end{tikzpicture}
\caption{}
\end{subfigure}

\begin{subfigure}[b]{0.45\textwidth}
\centering
\begin{tikzpicture}
  \node (img)  {\includegraphics[width=0.8\textwidth]{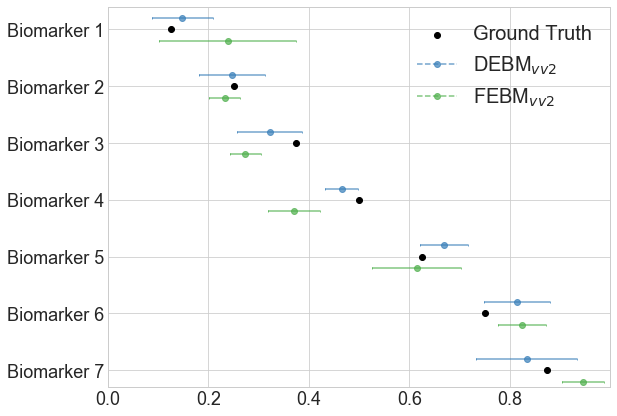}};
  \node[below=of img, node distance=0cm, yshift=1.2cm,font=\color{black}] {\small Disease Stage};
\end{tikzpicture}
\caption{}
\end{subfigure}
\caption{Experiment 4: Figures (a) and (b) show the event-center errors of DEBM\textsubscript{vv2} and FEBM\textsubscript{vv2} as a function of $\Sigma_\beta$ and $\Sigma_\xi$ respectively. Figure (c) shows the estimated event-center locations for both methods as well as the ground-truth event centers. Error bars in (a), (b) and (c) represent standard deviation over $50$ repetitions of simulation.}
\label{fig:VaryNoise2}
\end{figure}
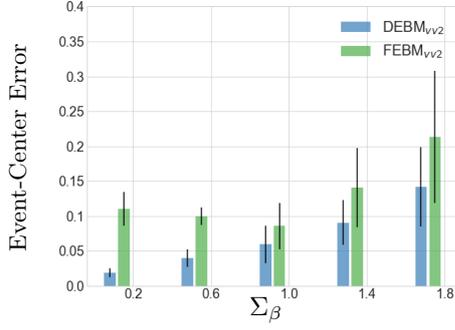
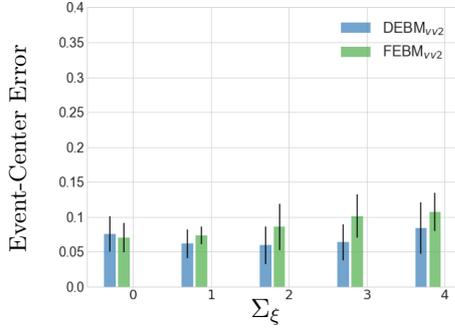
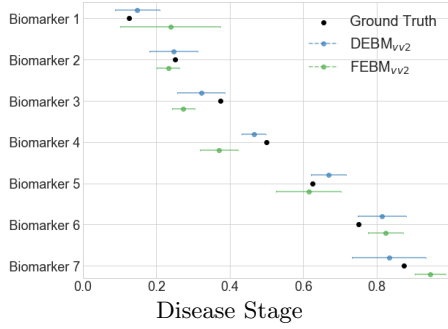

\textbf{Experiment 4:} Figure~\ref{fig:VaryNoise2} (a) and (b) shows the event-center errors in DEBM\textsubscript{vv2} and FEBM\textsubscript{vv2} as the variability in population $(\Sigma_\beta)$ and disease heterogeneity $(\Sigma_\xi)$ increases respectively. It should be noted from Figure~\ref{fig:VaryNoise1}(b) and Figure~\ref{fig:VaryNoise2} (b) that, even when the FEBM\textsubscript{vv2} gets the ordering more accurately than DEBM\textsubscript{vv2} in cases of low $\Sigma_\xi$, the event-center estimation of DEBM\textsubscript{vv2} is on par with or better than its FEBM counterpart.

Figure~\ref{fig:VaryNoise2} (c) shows the estimated event-center locations for $\Sigma_\beta = 1.0$ and $\Sigma_\xi = 2$ and the ground truth event-centers.

\begin{figure}[t!]
    \centering
    \begin{subfigure}[b]{0.45\textwidth}
    \centering
\begin{tikzpicture}
  \node (img)  {\includegraphics[width=0.85\textwidth]{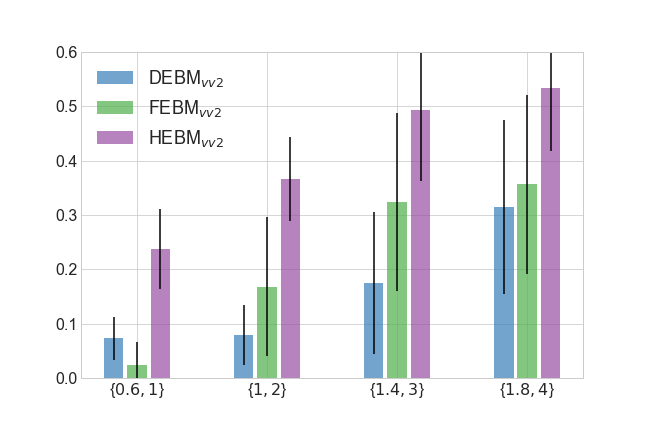}};
  \node[below=of img, node distance=0cm, yshift=1.5cm,font=\color{black}] {$\{ \Sigma_\beta, \Sigma_\xi \}$};
  \node[left=of img, node distance=0cm, rotate=90, anchor=center,yshift=-1cm,font=\color{black}] {\small Ordering Error};
 \end{tikzpicture}
 \caption{}
\end{subfigure}

\begin{subfigure}[b]{0.45\textwidth}
\centering
\begin{tikzpicture}
  \node (img)  {\includegraphics[width=0.8\textwidth]{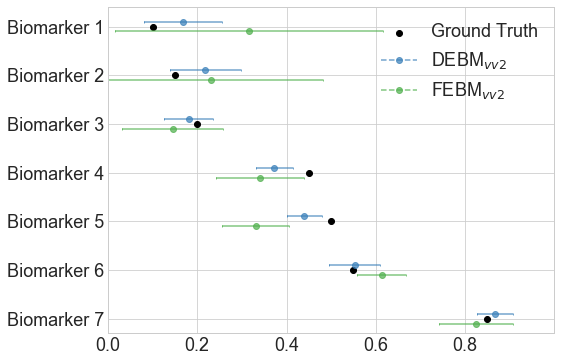}};
  \node[below=of img, node distance=0cm, yshift=1.3cm,font=\color{black}] {Disease Stage};
  \node[left=of img, node distance=0cm, rotate=90, anchor=center,yshift=-1cm,font=\color{black}] {};
\end{tikzpicture}
\caption{}
\end{subfigure}

\begin{subfigure}[b]{0.45\textwidth}
\centering
\begin{tikzpicture}
  \node (img)  {\includegraphics[width=0.85\textwidth]{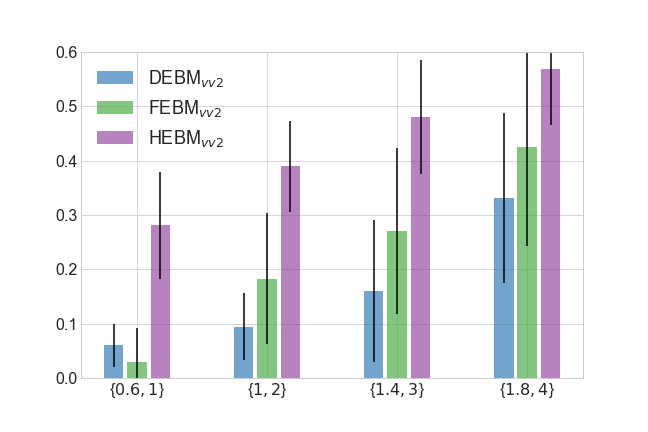}};
  \node[below=of img, node distance=0cm, yshift=1.5cm,font=\color{black}] {$\{ \Sigma_\beta, \Sigma_\xi \}$};
  \node[left=of img, node distance=0cm, rotate=90, anchor=center,yshift=-1cm,font=\color{black}] {\small Ordering Error};
\end{tikzpicture}
\caption{}
\end{subfigure}


\caption{Experiment 5: Figures (a) and (c) show the ordering errors of DEBM\textsubscript{vv2}, FEBM\textsubscript{vv2} and HEBM\textsubscript{vv2} when  $\mu_{\xi_i}$ are not uniformly distributed. $\Sigma_{\beta}$ and $\Sigma_{\xi}$ increase as we move from left to right. Figure (a) shows the errors in the case when $\rho_{i}$ are identical for all the biomarkers whereas (c) shows the errors when $\rho_{i}$ are different. Figure (b) shows the non-uniform $\mu_{\xi_i}$ as well as the estimated event-centers by DEBM\textsubscript{vv2} and FEBM\textsubscript{vv2} for the case of $\rho_{i}$ being equal. Error bars in (a), (b) and (c) represent standard deviation over $50$ repetitions of simulation.}
\label{fig:VarySpacing}
\end{figure}
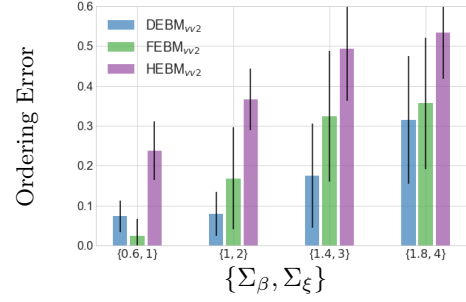
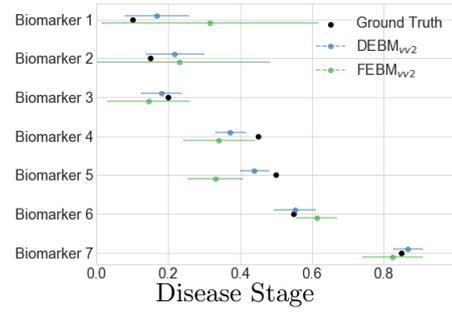
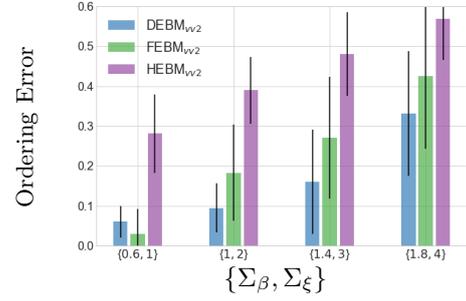

\textbf{Experiment 5:} Figure~\ref{fig:VarySpacing} (a) shows the ordering errors of DEBM\textsubscript{vv2}, FEBM\textsubscript{vv2} and HEBM\textsubscript{vv2} as $\Sigma_\beta$ and $\Sigma_\xi$ increase, when the ground-truth event centers ($\mu_{\xi_i}$) are non-uniformly spaced. The spacing of $\mu_{\xi_i}$ can be observed in Figure~\ref{fig:VarySpacing} (b), where the ground truth event-centers as well as the estimated event-centers of DEBM\textsubscript{vv2} and FEBM\textsubscript{vv2} are shown for $\Sigma_\beta = 1.0$ and $\Sigma_\xi = 2$. It can be observed that the estimated event-centers for DEBM\textsubscript{vv2} are much closer to the ground-truth event centers than those of FEBM\textsubscript{vv2} and also have a much lower variance over different iterations of simulations.

Figure~\ref{fig:VarySpacing} (c) shows the ordering errors as $\Sigma_\beta$ and $\Sigma_\xi$ increases, when $\mu_{\xi_i}$ is non-uniformly spaced and $\rho_{i}$ is not identical for all biomarkers. It should also be noted that the mean of $\rho_{i}$ over all $i$ has not changed between (a) and (c). The variation of errors in (c) is quite similar to the one in (a). This shows that performance of EBM methods that are reported in other experiments (where $\rho_{i}$ is equal for all biomarkers) can be expected to not deteriorate in the more realistic scenario of $\rho_{i}$ not being equal for all biomarkers. The event-center variance for $\Sigma_\beta = 1.0$ and $\Sigma_\xi = 2$ for the case of unequal $\rho_{i}$ is very similar to (b) and has been included as supplementary material (Figure S2).

\begin{figure}[t!]
\centering
\begin{tikzpicture}
  \node (img)  {\includegraphics[width=0.45\textwidth]{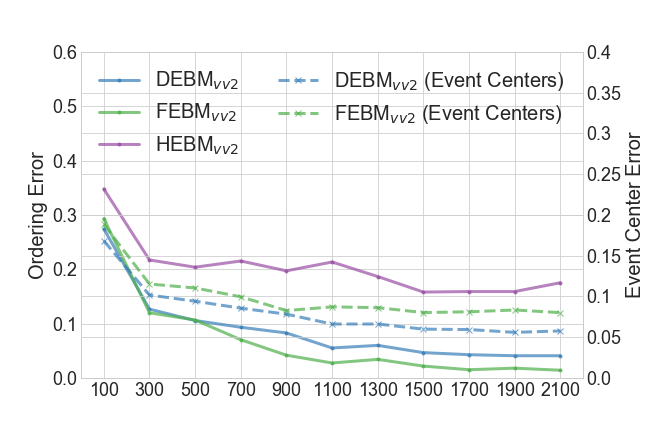}};
  \node[below=of img, node distance=0cm, yshift=1.3cm,font=\color{black}] {Number of Subjects (M)};
\end{tikzpicture}
\caption{Experiment 6: Ordering errors of DEBM\textsubscript{vv2}, FEBM\textsubscript{vv2} and HEBM\textsubscript{vv2} as a function of number of subjects $(M)$ in the dataset. It also shows the event-center errors of DEBM\textsubscript{vv2} and FEBM\textsubscript{vv2} as a function of $M$.}
\label{fig:VarySubjects}
\end{figure}

\textbf{Experiment 6:} Figure~\ref{fig:VarySubjects} shows the mean ordering errors of DEBM\textsubscript{vv2}, FEBM\textsubscript{vv2} and HEBM\textsubscript{vv2} as a function of number of subjects in the dataset on one vertical axis and shows the mean event-center errors of DEBM\textsubscript{vv2} and FEBM\textsubscript{vv2} on the other vertical axis. As expected, the models perform better as the number of subjects increases. DEBM\textsubscript{vv2} is slightly better at inferring the central ordering than FEBM\textsubscript{vv2} when the number of subjects is very low, but FEBM\textsubscript{vv2} outperforms DEBM\textsubscript{vv2} when the number of subjects is higher. However, when the accuracy of event centers are considered, DEBM\textsubscript{vv2} consistently outperforms FEBM\textsubscript{vv2}.

\begin{figure}[t!]
\centering
\begin{tikzpicture}
  \node (img)  {\includegraphics[width=0.45\textwidth]{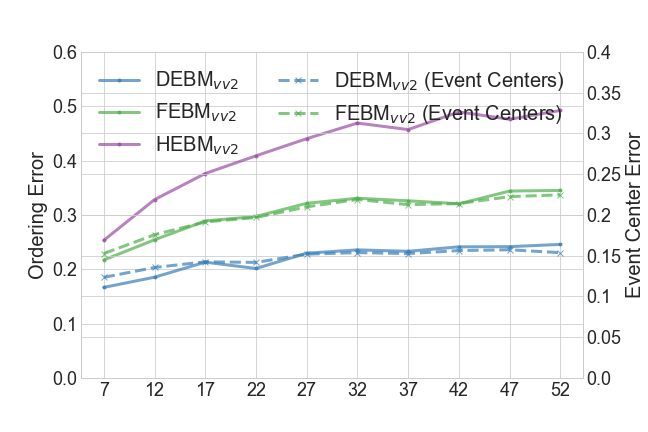}};
  \node[below=of img, node distance=0cm, yshift=1.3cm,font=\color{black}] {Number of Events (N)};
\end{tikzpicture}
\caption{Experiment 7: Ordering errors of DEBM\textsubscript{vv2}, FEBM\textsubscript{vv2} and HEBM\textsubscript{vv2} as a function of number of events $(N)$ in the dataset. It also shows the event-center errors of DEBM\textsubscript{vv2} and FEBM\textsubscript{vv2} as a function of $N$.}
\label{fig:VaryEvents}
\end{figure}

\textbf{Experiment 7:} Figure~\ref{fig:VaryEvents} shows the mean ordering errors of DEBM\textsubscript{vv2}, FEBM\textsubscript{vv2} and HEBM\textsubscript{vv2} as a function of the number of events (biomarkers) in the dataset on one vertical axis and shows the mean event-center errors of DEBM\textsubscript{vv2} and FEBM\textsubscript{vv2} on the other vertical axis. The biomarkers were selected randomly after replacement so that the chances of selecting a bad biomarker remain equal as the number of events increases. It can be noted that the errors of the EBM models increase as the number of events increases initially, even when the average quality of biomarkers remains the same. However the errors stabilize beyond a certain point and do not increase any more.

\section{Discussion} \label{sec:disc}

We proposed a novel discriminative EBM framework to estimate the ordering in which biomarkers become abnormal during disease progression, based on a cross-sectional dataset. The proposed framework outperforms state-of-the-art EBM techniques in estimating the event ordering. We also introduced the concept of relative distance between event-centers, which enables creating a disease progression timeline. This in turn led to the development of a new continuous patient staging mechanism. In addition to the framework, we also proposed a novel probabilistic Kendall's Tau distance metric and a robust biomarker distribution estimation algorithm. In this section, we discuss different aspects of the proposed algorithm.

\subsection{Event Centers}

Event-centers capture relative distance between events. This helps in creating the disease progression timeline from an ordering of events. Event centers are an intrinsic property of the biomarker used, for the selected population. This was observed in Experiment $1$(b) where the event-centers estimated using DEBM\textsubscript{vv2} remained fairly consistent $(\pm 0.05)$ across models using different number of biomarkers.

The estimated disease progression timeline can be used for inferring progression of the disease, with the event centers being synonymous to milestones of progression. A strict quantization of position in ordering of events (as reported in~\cite{Neil:2017}, ~\cite{Vikram:2017}, ~\cite{Young:2015}, ~\cite{Young:2014}, ~\cite{Fonteijn:2012}) in the positional variance diagram can sometimes be non-intuitive in terms of inferring actual progression of the disease. This was seen in Experiment $1$(a), where the event center variance diagram showed that the TAU event (at position 6) was closer to the p-TAU event (at position 2) than the whole brain event (position 7). 

The approach of scaling the event-centers between $\left[0,1\right]$ has its advantages and disadvantages. The advantage of such a scaling is that models built on different biomarkers, but within the same population, remain comparable. For example, a model built with CSF and MRI based biomarkers can be compared with a model built on MRI based biomarkers alone, as the event-centers of MRI based biomarkers would approximately be the same. On the other hand, the position of the first event relies heavily on the number of `true' controls in the dataset (CN subjects who are not in an early asymptomatic stage of the disease). This is the result of introducing pseudo-events for scaling the events-centers. 

Comparison of the event centers across different datasets with different number of controls (albeit with the same biomarkers) can be done in three ways. Event-centers can be scaled and translated such that the mean and standard deviation of event centers computed across different datasets are the same (similar to the comparisons between estimated and ground-truth event centers in this paper). Alternately, the event center of the first biomarker can be set as $0$ and the event center of the last biomarker can be set to $1$, before comparison. Lastly, in a dataset where controls (i.e., subjects whose biomarker values are all normal) can be easily identified, it would be better to exclude them for event-center computation.

The estimated event centers have a good correlation with the groundtruth disease timeline. This can be seen in the simulation experiments with and without uniform spacing of events (experiments $4$ and $5$). It must however be noted that, the disease stages $\Psi$ of the simulated subjects were distributed uniformly throughout the disease timeline. If the distribution is not uniform, we expect it to have an effect on the estimation of event centers. Analyzing the exact effect of such non-uniform distributions on the estimation of event centers and ways to estimate event centers invariant to the distribution of subjects on the disease timeline could be an interesting extension of the current work.

Experiment $5$ also showed that different biomarkers having different rates of progression does not degrade the performance of EBM models, as long as the mean rate of progression is the same. We did not perform an experiment to benchmark the accuracies by changing the mean rates of progression of biomarkers. This experiment was already performed in~\cite{Young:2015b} and it was observed that FEBM ordering error decreases as the mean rates of progression increase.

FEBM assumes that the disease is homogeneous, as it expects all the subjects in the dataset to follow the same ordering. When the variability of ordering in different subjects is low, FEBM with the proposed mixture model `vv2' outperforms DEBM with the proposed mixture model. This can be seen in the results of Experiments $3$, $5$ and $6$. When the assumption becomes too restrictive, DEBM with the proposed mixture model outperforms FEBM. Even when the assumption holds true, estimation of event-centers with DEBM is more accurate than with FEBM.

\subsection{Patient Staging}

Existing patient staging algorithms discretize the patient stages based on event position, whereas the patient staging algorithm introduced in this paper takes relative distance between events into consideration while staging new subjects. This makes patient stages more useful for diagnosis and prognosis as they correlate more with the actual disease progression timeline. Discrete patient stages without considering the event centers could diminish the prognosis value of the obtained stages.

The cross-validation experiment on ADNI data (Experiment $2$) showed that the CN and AD subjects are well separated after patient staging and that the AUC of the proposed method is better than that of the state-of-the-art EBM techniques. It also showed that MCI converters and non-converters are well separated after patient staging, without explicitly training the model to achieve this.

It must however be noted that even though heterogeneity of the disease was considered while inferring the central ordering, it was not considered for patient staging. Inferring multiple central orderings corresponding to different disease subtypes~\citep{Young:2015} and staging patients on one of these central orderings may help us overcome this drawback. Patient staging with respect to subject-specific orderings (as done in HEBM) can also be considered when extending DEBM for longitudinal data, where the subject-specific orderings might be estimated with higher confidence.


\subsection{Scalability of Event-Based Models}

Understanding the progression of several imaging and non-imaging biomarkers after disease onset is important for assessing the severity of the disease. Hence it is desirable to have a model scalable to a large number of biomarkers. FEBM and DEBM are scalable to large number of events, whereas HEBM is not. This was seen in the simulation experiment on varying number of events (Experiment $7$), where the errors of FEBM and DEBM increased asymptotically with increasing number of events. The ordering errors of HEBM reached $0.5$ for large number of events, which is equivalent to random prediction.

In Experiment $6$, we observed that the errors of the EBMs decrease with increasing number of subjects in the dataset. We hence expect FEBM, DEBM and HEBM\textsubscript{vv2} to be scalable to a large number of subjects.

The performance of HEBM\textsubscript{jh} is seen to be consistently worse than FEBM\textsubscript{ay} in Experiment $3$. This is in contrast with the findings of~\cite{Vikram:2017}, where HEBM\textsubscript{jh} performed better than FEBM\textsubscript{ay} when the number of biomarkers used were $7$, while it performed worse when the number of biomarkers used were $42$. One of the key differences between the experiment performed in~\cite{Vikram:2017} and Experiment $3$ is the number of subjects in the simulation dataset. While the previous study considered $509$ subjects, Experiment $3$ considered $1737$ subjects. HEBM\textsubscript{jh} jointly estimates the subject-specific orderings of all the subjects and the mixture model to represent the biomarkers in different diagnostic groups. We think that while the joint estimation was good for low number of subjects, increasing the number of subjects had an adverse effect on the convergence of the algorithm. Hence HEBM\textsubscript{jh} is not scalable to a large number of subjects.

We decoupled the mixture model and estimation of subject-specific orderings in HEBM\textsubscript{vv2} (Experiments $3$, $5$, $6$ and $7$). This made HEBM more scalable as it improved the results in Experiment $3$ with $1737$ subjects, but the decoupling had an adverse effect on the algorithm when the number of subjects was low, as seen in Experiment $6$, where HEBM\textsubscript{vv2} performs worse than FEBM\textsubscript{vv2} even when the number of subjects was low.

FEBM and HEBM are generative approaches for estimating the central ordering. Our results suggest that HEBM is not very scalable. Although FEBM is scalable, the assumptions made in FEBM are too restrictive for heterogeneous disease such as AD. DEBM is a discriminative approach to event-based modeling, which is both scalable and can robustly estimate central ordering even when the disease is heterogeneous.

\subsection{The Mixture Model}

The optimization technique for the Gaussian mixture model that is presented in this paper decouples the optimization of Gaussian parameters and mixing parameters. When the Gaussians of the pre-event and post-event classes are highly overlapping, the optimum mixing parameter changes a lot even for small changes in the Gaussian parameters. By decoupling the optimizations for Gaussian parameters and mixing parameters, we get more stable mixing parameters. This helps in improving the accuracy of all EBMs. This was observed in Experiment $3$.

\subsection{The Importance of Good Biomarkers}

Quality of biomarkers plays a huge role in the accuracy of the EBMs. This was seen in Experiment $7$, where the mean error value for $7$ biomarkers was considerably higher than the mean error value with the same number of biomarkers in Experiment $4$ (for the same $\Sigma_\beta$ and $\Sigma_\xi$ parameters). The observed difference can be explained by the choice of the biomarkers used in those experiments. While the biomarkers chosen in Experiment $7$ was at random, the ones chosen in Experiment $4$ were the $7$ best biomarkers.

\subsection{Interpretation of model results on ADNI}

Experiment $1(a)$ showed that CSF biomarkers ABETA and p-TAU are one of the first biomarkers that become abnormal in AD, which is in agreement with Jack's hypothetical model~\citep{Jack:2013}. The event centers of Hippocampus volume and TAU are quite close to each other as well, which is also in agreement with the current understanding of the disease~\citep{deSouza:2012}. However, MMSE and ADAS biomarkers are seen to become abnormal quite early in the disease as well, which is not in agreement with the hypothetical model. This could suggest that functional abnormality starts before structural abnormality. It could also be because of the fact that we included everyone in ADNI baseline measurement, for estimating the ordering. Subjects with MCI may not necessarily be progressing towards AD and including these subjects could have an effect on inferring the ordering. 

Nucleus accumbens right and left are the first biomarkers to become abnormal as seen in Figure $7$. This was also observed by~\cite{Young:2017}. However, the large standard error of the event centers for the events before ABETA suggests that the exact position of those events are unreliable.  Experiment $1(b)$ showed that weak biomarkers (biomarkers excluded in Figure~\ref{fig:ADNI25}, but included in Figure~\ref{fig:ADNI50}) could lead to greater uncertainty in event centers. This can be explained by the fact that weak biomarkers are the ones where there is a lot of overlap between the Gaussians of pre-event and post-event classes. Small variation in the sampling population during bootstrapping leads to large changes in the parameters estimated in the mixture modeling step of the algorithm. It also showed that majority of the early structural biomarkers are from Temporal lobe, followed by Central structures, Frontal lobe, Parietal lobe and Occipital lobe.

\section{Conclusion} \label{sec:conc}

We proposed a new framework for event-based modeling, called discriminative event-based modeling (DEBM), which includes a new optimization strategy for Gaussian mixture modeling, a new paradigm for inferring the mean ordering, a way for estimating the proximity of events in the order to create a disease progression timeline, and a new way of staging patients that uses these relative proximities of events while placing new subjects on the estimated timeline. The source code for DEBM and FEBM was made publicly available online under the GPL 3.0 license: \url{https://github.com/88vikram/pyebm/}.

We applied the DEBM framework to a set of $1737$ subjects from the baseline ADNI measurement, and also performed an extensive set of simulation experiments verifying the technical validity of DEBM. The experiment on ADNI data illustrated a number of advantages of the new approach. Firstly, we showed that strict quantization of position in ordering of events in the positional variance diagram can sometimes be non-intuitive in terms of inferring actual progression of a disease. Secondly, we showed that the patient staging based on the proposed approach separates CN and AD group of subjects much better than the previous EBM models. Thirdly, we showed that the patient staging can be used to identify individuals at-risk of developing AD as the MCI converters and non-converters were well-separated. Staging patients based on the estimated disease progression timeline can thus make computer-aided diagnosis and prognosis more explainable. The results of these experiments are encouraging and suggest that DEBM is a promising approach to disease progression modeling.

\section*{Acknowledgement}

This work is part of the EuroPOND initiative, which is funded by the European Union's Horizon 2020 research and innovation programme under grant agreement No. 666992. The authors thank Dr. Jonathan Huang for sharing the implementation of Huang's EBM and Dr. Alexandra Young for sharing the implementation of the simulation system for biomarker evolution.

Data collection and sharing for this project was funded by the Alzheimer's Disease Neuroimaging Initiative (ADNI) (National Institutes of Health Grant U01 AG024904) and DOD ADNI (Department of Defense award number W81XWH-12-2-0012). ADNI is funded by the National Institute on Aging, the National Institute of Biomedical Imaging and Bioengineering, and through generous contributions from the following: AbbVie, Alzheimer’s Association; Alzheimer's Drug Discovery Foundation; Araclon Biotech; BioClinica, Inc.; Biogen; Bristol-Myers Squibb Company; CereSpir, Inc.; Cogstate; Eisai Inc.; Elan Pharmaceuticals, Inc.; Eli Lilly and Company; EuroImmun; F. Hoffmann-La Roche Ltd and its affiliated company Genentech, Inc.; Fujirebio; GE Healthcare; IXICO Ltd.; Janssen Alzheimer Immunotherapy Research \& Development, LLC.; Johnson \& Johnson Pharmaceutical Research \& Development LLC.; Lumosity; Lundbeck; Merck \& Co., Inc.; Meso Scale Diagnostics, LLC.; NeuroRx Research; Neurotrack Technologies; Novartis Pharmaceuticals Corporation; Pfizer Inc.; Piramal Imaging; Servier; Takeda Pharmaceutical Company; and Transition Therapeutics. The Canadian Institutes of Health Research is providing funds to support ADNI clinical sites in Canada. Private sector contributions are facilitated by the Foundation for the National Institutes of Health (www.fnih.org). The grantee organization is the Northern California Institute for Research and Education, and the study is coordinated by the Alzheimer's Therapeutic Research Institute at the University of Southern California. ADNI data are disseminated by the Laboratory for Neuro Imaging at the University of Southern California. 




\bibliographystyle{model2-names.bst}\biboptions{authoryear}





\nocite{*}

\section*{\refname}
\bibliography{EBM}
\linenumbers
\end{document}